\documentclass[final,3p,times]{elsarticle}

\makeatletter
\def\ps@pprintTitle{%
  \let\@oddhead\@empty
  \let\@evenhead\@empty
  \def\@oddfoot{\reset@font\hfil}%
  \let\@evenfoot\@oddfoot
}
\makeatother

\usepackage{setspace}

\usepackage[dvipsnames,svgnames,x11names]{xcolor}
\usepackage[draft]{changes}
\usepackage[inline]{enumitem}
\definechangesauthor[color=blue]{PC}
\usepackage{subcaption}
\usepackage{lineno,hyperref}
\usepackage{amsmath,amssymb,amsfonts}
\usepackage{algorithmic}
\usepackage{graphicx}
\usepackage{anyfontsize}
\usepackage{adjustbox}
\usepackage{rotating}
\usepackage{tablefootnote}
\usepackage{threeparttable}
\usepackage{longtable}
\usepackage{tabularray}
\usepackage{caption}
\usepackage[font=small,labelfont=bf, figurename=Fig.]{caption} 

\usepackage{textcomp}
\usepackage{todonotes}
\usepackage{tabularx}
\usepackage{gensymb}
\usepackage[utf8]{inputenc}
\usepackage{mathtools, nccmath}
\usepackage{booktabs}
\usepackage{cleveref}
\usepackage{chemformula}
\usepackage[version=4]{mhchem}
\usepackage{booktabs, multirow} % for borders and merged ranges
\usepackage{soul}% for underlines
\modulolinenumbers[1]

\biboptions{authoryear}

\usepackage{comment}
\usepackage[inline]{enumitem}   
\makeatletter
\newcommand{\inlineitem}[1][]{%
\ifnum\enit@type=\tw@
    {\descriptionlabel{#1}}
  \hspace{\labelsep}%
\else
  \ifnum\enit@type=\z@
       \refstepcounter{\@listctr}\fi
    \quad\@itemlabel
    \hspace{\labelsep}%
\fi}
\makeatother
\begin{document}

\begin{frontmatter}

%\title{Traffic Entity based Driver Gaze Estimation using Transformer Architecture}
%\title{Driver Gaze Object Prediction Using Transformer-Based Cross Attention of Face and Scene Features}

%\title{TransGaze-Object: Transformer Gaze Object Estimation Framework for Driver Gaze Estimation}

%\title{TransGaze-Object: A Transformer-Based Gaze Object Prediction Framework for Driver Gaze}
%\title{A Transformer-Based Framework for Driver Gaze-Object Prediction in Real-World Driving}

\title {TransGaze-Object: Transformer Based Driver Gaze Object Prediction Framework in Real Driving}
%\titel {A Transformer-Based Framework for Driver Gaze Object Prediction Using a Heterogeneous Real-World Driving Dataset}

% \address{Radarweg 29, Amsterdam}


\author[mymainaddress]{Pavan Kumar Sharma}

\author[mymainaddress1]{Ayush Pande}

% \author[mymainaddress2]{Arvind Kumar}

% \author[mymainaddress]{Abhay Shukla}

\author[mymainaddress]{Pranamesh Chakraborty\corref{mycorrespondingauthor}}

\cortext[mycorrespondingauthor]{Corresponding author\\
Pavan Kumar Sharma: pavans20@iitk.ac.in, Ayush Pande: ayushp@cse.iitk.ac.in, Pranamesh Chakraborty: pranames@iitk.ac.in, +91-512-259-2146}
% %\ead{support@elsevier.com}

\address[mymainaddress]{Department of Civil Engineering, Indian Institute of Technology Kanpur, Kanpur-208016, U.P., India}

\address[mymainaddress1]{Department of Computer Science and Engineering, Indian Institute of Technology Kanpur, Kanpur-208016, U.P., India}

% \address[mymainaddress2]{SAP Labs India, Bengaluru–560066, Karnataka, India}

\begin{abstract}

Driver gaze provides information regarding driver visual attention and situational awareness  to the surrounding traffic. Existing driver gaze estimation studies represent gaze in terms of gaze zone or gaze vector/point-of-gaze (PoG). However, object-level gaze information provides a more semantically meaningful representation of visual attention by identifying attended objects, such as vehicles, pedestrians, or traffic signals. In this study, we propose an end-to-end driver gaze object prediction framework, \textbf{\textit{TransGaze-Object}}, \textbf{T}ransformer-based \textbf{G}aze \textbf{O}bject prediction model. The proposed framework first extracts facial features, including face and iris-weighted eye features, along with traffic-object spatial features. A transformer based cross-attention mechanism is then used to compute similarity scores and attention weights for predicting the driver's gaze object. To train this model, we propose a benchmark driver gaze dataset, \textbf{U}rban \textbf{D}riving\textbf{-F}ace \textbf{S}cene \textbf{G}aze (\textbf{UD-FSG}), comprising synchronized driver-face and traffic-scene images, scene objects bounding boxes, and gaze labels in terms of 2D gaze coordinate and gaze object. The \textit{TransGaze-Object} model achieves an overall accuracy of 60\% for gaze-object prediction, compared to 51\% accuracy obtained from associating the estimated Point-of-Gaze to traffic objects. 
The error analysis reveals that \textit{TransGaze-Object} reduces confusion between traffic objects (predicted) and the background (ground-truth), achieving an error rate of 11.68\%, a 49.7\% relative reduction compared with 23.21\% error obtained from PoG-based gaze-object association. Overall, the results demonstrate the effectiveness of directly predicting gaze objects from driver-face and traffic-scene information, rather than estimating an intermediate Point-of-Gaze and subsequently associating it with traffic objects.
\end{abstract}
\begin{keyword}
Driver Gaze Estimation \sep Driver Visual Attention \sep Transformer \sep Gaze Object Prediction %\sep Driver Gaze Dataset  
\end{keyword}

\end{frontmatter}
\section{Introduction}
 Driver safety has been a major global concern for decades due to the large number of road accidents occurring every year. According to the WHO Global Status Report on Road Safety \citep{WHO2023RoadSafety}, approximately 1.19 million people died in road crashes in 2023. These road crashes occur due to several reasons and can be broadly categorized into four major groups: human factors, vehicle defects, road conditions, and environmental factors. Among these, human factors are considered a leading cause of road crashes. Crashes associated with human factors, particularly driver-related factors, may arise from fatigue, distracted driving, variations in driver cognitive states, visual attention, and situational awareness \citep{lal2001critical, regan2008driver, dingus2016driver, louw2017you, li2019drivers, li2023much}. These factors can impair drivers' ability to perceive and respond appropriately to changes in the driving environment, potentially increasing the risk of crashes. However, directly measuring some of these human-related factors, such as visual attention and situational awareness, is challenging. In this context, the driver's gaze plays a significant role as an alternative measure of visual attention and situational awareness. Apart from this, driver gaze is also used for several other important applications, including driver monitoring systems and the development of advanced driver assistance systems (ADAS).
\par Driver gaze estimation refers to the process of determining where a driver is looking while driving. Typically, the output of gaze estimation models can be represented in terms of gaze zone, gaze vector/point of gaze (PoG), or gaze object \citep{sharma2024review}. Gaze-zone based representation divides the windshield and surrounding areas (e.g., side mirrors, rear view mirror, ceterstack, etc.) into different zones, and driver gaze zone classification involves estimating which zone the driver is looking at \citep{fridman2016driver, martin2018dynamics, ghosh2021speak2label, wu2025multi, yahyaabadi2026driver}. Driver gaze zone representation does not inherently account for traffic scene information. On the other hand, point of gaze-based representation involves estimating the corresponding point on the windshield \citep{vicente2015driver, yuan2022self, cheng2024you} or the traffic scene image the driver is looking at \citep{kasahara2022look, hu2025lnet, zhou2025eraw}. However, the driver's gaze point alone does not provide information about the driver's attention towards the objects (vehicles, pedestrians, etc.) in the traffic scene. Therefore, this necessitates post-processing the PoG information to determine the traffic object (if any) the driver is looking at. In contrast, driver gaze object representation involves determining the gaze object the driver is looking at.
\par Existing studies on driver gaze estimation have been based on gaze zone classification or PoG estimation \citep{chuang2014estimating, tawari2014driver, fridman2016owl, vora2018driver, yang2019dual, yuan2022self, kasahara2022look, wu2025multi, li2026geometry}. To the author's knowledge, there is currently no study that has worked on driver gaze object prediction. We argue that providing the information of the objects in the traffic scene (\textbf{apriori}) along with the driver face information, and converting the problem of gaze estimation to an end-to-end gaze object prediction problem, can help to improve the estimation results, compared to PoG estimation first and then post-processing to obtain the driver gaze object information. Therefore, in this study, we propose driver gaze estimation as an end-to-end driver gaze object prediction framework, which takes as input the driver's face and the traffic objects (vehicles, pedestrians, traffic signs, etc.) and predicts which object among the given objects (or the background) the driver is gazing at.
% In fact, literature also shows that driver attention is object-centered \citep{duncan1984selective, driver1989movement, land1994we, watson1999object, underwood2007visual}. Therefore, it makes sense to use object information directly for estimating gaze. 
% Object-based gaze estimation approach offers a more informative and semantically meaningful description of driver visual attention compared to zone-based and point-of-gaze (PoG) representations. Object-based gaze directly associates the driver's gaze with scene entities such as vehicles, pedestrians, or traffic signals, which aligns more closely with driving cognition and decision-making. Therefore in this study, we propose driver gaze estimation as an end-to-end driver gaze object prediction framework which takes as input the driver face and the traffic objects (vehicles, pedestrians, traffic signs, etc.) and predicts which object among the given objects (or the background) the driver is gazing.
\par Driver gaze object prediction requires fusing the information from two different modalities: (i) driver face information, which can be extracted from a driver face image captured by a camera facing the driver, and (ii) traffic objects information, which can be captured by a camera facing the road. In this study, we propose to fuse this information with a cross-attention mechanism \citep{vaswani2017attention}. Here, the driver's facial features can be taken as query vectors, and the traffic object features (object bounding box size and location in the traffic scene) as key vectors. This query-key representation of driver faces and traffic objects, and their fusion using a cross-attention mechanism, helps understand the relationship between face features and traffic objects, determining which object in the traffic scene (or background) the driver is gazing at.
\par Driver gaze object prediction model training requires a dataset comprising synchronized driver face images and traffic scene images. However, existing driver gaze datasets typically consist of only driver face images \citep{ribeiro2019driver, rangesh2020driver, dua2020dgaze, ghosh2021speak2label, sharma2025evaluation}. This is because existing studies outputs are in terms of gaze zones or gaze vectors (an alternative representation of PoG), which do not require traffic scene information. Currently, to our knowledge, there exists only one open-source benchmark driver gaze dataset, Look Both Ways (LBW) \citep{kasahara2022look}, which contains both driver face and traffic scene images. However, the LBW dataset was collected mostly in low-density traffic, and the number of traffic objects in the scene is very low (an average of 4 objects per image). Therefore, in this study, we develop a driver gaze dataset comprising synchronized pairs of driver face and scene images, scene traffic object bounding box coordinates, and gaze labels expressed as 2D gaze coordinates and corresponding gaze-object information. The data was collected under high-density, heterogeneous urban traffic conditions, which make the gaze object prediction problem challenging due to a larger number of potential gaze objects (i.e., traffic objects) in the traffic scene.
\par The major contributions of this study are as follows:
%\begin{itemize}
\begin{enumerate}[label=(\roman*)]
    %\item We propose driver gaze estimation as an end-to-end gaze-object prediction problem, taking the driver's face image and object information in the traffic scene as inputs directly and predicting the object the driver is gazing at.
    \item We formulate driver gaze estimation as an end-to-end gaze-object prediction problem that directly takes the driver’s face image and traffic-scene object information as inputs and predicts the object toward which the driver is gazing.  
    \item We develop a novel Transformer architecture-based gaze-object prediction model (TransGaze-Object), in which face input features are integrated with traffic object information via a cross-attention mechanism.
    % \item We develop a dataset called UD-FSG, collected in real-world urban driving environments, which comprises synchronized driver face and traffic scene images, bounding box information for traffic objects, corresponding gaze ground truth in terms of gaze-object labels, and 2D gaze points relative to the traffic scene image.
    \item We develop a dataset called UD-FSG, collected in real-world urban driving scenarios, which comprising synchronized pairs of driver face and scene images, scene traffic object bounding box coordinates, and gaze labels expressed as 2D gaze coordinates and corresponding gaze-object information.
\end{enumerate}
%\end{itemize}

\par The remainder of this paper is organized as follows. Section 2 provides a detailed review of driver gaze estimation, specifically recent developments in point-of-gaze and gaze-vector-based studies, including transformer-based architectures. Section 3 describes the datasets, which include sensor setup, data collection, ground truth creation, and a benchmark gaze-object-based dataset. Section 4 presents the proposed methodology, which includes face and scene object detection and feature extraction; feature fusion and computation of attention weights to predict the gaze object. Section 5 presents the experimental results, including training performance, model evaluation, a comparison of our proposed model's performance with PoG-based object association, and an error analysis comparing predicted and ground-truth gaze objects to identify possible reasons for incorrect gaze object predictions. At last, we conclude key findings, strengths, limitations of our proposed model, and the future scope of the study.

\section{Literature review}
\par Intrusive and non-intrusive are the two approaches of driver gaze estimation, based on the device used. In the intrusive method, drivers wear a head-mounted device, also referred to as an eye tracker.  
In contrast, non-intrusive approaches to driver gaze estimation rely on cameras mounted on the vehicle's dashboard and/or windshield to capture the driver's face \citep{ortega2020dmd, ghosh2021speak2label, sharma2025evaluation}. Since, in a real, practical driver monitoring system, the driver's gaze needs to be continuously monitored to assist and ensure driver safety, a non-intrusive approach is more suitable. In this section, we only review non-intrusive approach-based gaze estimation studies.
% \par Gaze estimation in in-vehicle environments has traditionally focused on representing the driver's gaze as a predefined gaze zone corresponding to the region toward which the driver is looking. Several studies have investigated this zone-based gaze estimation approach \citep{fridman2016driver, martin2018dynamics, yang2019dual, yang2021driver, shah2022driver, sharma2024driver, yahyaabadi2026driver, wang2026pigaze}. However, a major limitation of zone-based gaze estimation methods is that they do not provide information about the objects within the scene on which driver is focusing and are primarily limited to predefined regions of the vehicle interior, such as the side-view mirrors, windshield region, and rear-view mirror. 

% \par The gaze representation is in terms of gaze vector or gaze direction \citep{lrd2022distraction, kasahara2022look, yang2019dual}, which uses facial orientation, eye position, and head pose to compute a gaze vector. Further, this gaze direction is used to create a scene saliency map on the scene image \citep{kasahara2022look, wu2025multi}. This approach of gaze representation offers finer gaze estimation and is also suitable for continuous gaze monitoring in real-world applications. In the object-based gaze representation,  the driver's gaze is represented in terms of the object, such as a bicycle, car, bus, truck, etc., at which the driver is actually looking while driving.
\subsection{Literature on gaze zone classification and point-of-gaze estimation}
Gaze estimation in in-vehicle environments has traditionally focused on representing the driver's gaze as a predefined gaze zone corresponding to the region toward which the driver is looking. Several studies have investigated this zone-based gaze estimation approach \citep{fridman2016driver, martin2018dynamics, yang2019dual, yang2021driver, shah2022driver, sharma2024driver, yahyaabadi2026driver, wang2026pigaze}. However, a major limitation of zone-based gaze estimation methods is that they do not provide information about the objects within the scene on which driver is focusing and are primarily limited to predefined regions of the vehicle interior, such as the side-view mirrors, windshield region, and rear-view mirror.
\par Therefore, in recent years, several studies have worked on scene-based gaze representations, in which the driver's gaze is mapped directly onto the scene image. These approaches typically represent gaze either as a gaze vector/gaze direction \citep{lrd2022distraction, kasahara2022look, yang2019dual} and point of gaze or as a visual saliency map \citep{kasahara2022look, wu2025multi} indicating the region of the scene that attracts the driver's attention. The gaze vector or gaze direction based gaze representation uses probabilistic and deep learning models, such as CNN \citep{wu2025multi} and transformer-based methods \citep{hu2025lnet}. Some recent studies used a transformer-based gaze regressor that leverages face and eye features, as well as head movements (yaw, pitch, roll), to estimate the gaze vector. In some studies, a visual saliency map maps this gaze direction on the driver's scene image \citep{hu2021data, kasahara2022look}. A study by \cite{kasahara2022look} developed a self-supervised-based algorithm to estimate the driver gaze. The model takes a driver's face image as input and outputs 3D gaze direction and visual saliency in the scene. This study used the ETH-XGaze model, based on ResNet-50, for gaze estimation and the Unisal (MNet V2-RNN-Decoder) for saliency estimation. 
\cite{hu2021data} utilized the SalGAN adversarial framework for saliency map estimation, incorporating an element-wise sigmoid to interpret each pixel as a probability. Their architecture combined bottleneck, multi-resolution, and transition modules, enabling both down-sampling and up-sampling to enrich high-resolution features with global context. While this design strengthens feature fusion for dual-view gaze and scene representation, the study's reliance on a driving simulator limits its applicability. Real-world variability in environmental conditions and facial features remains unaddressed, reducing the system's generalizability.

% \cite{yang2019dual} developed a dual camera-based driver gaze estimation model in which a front dashboard camera captures the driver's face, and a camera mounted on the vehicle's interior roof captures the vehicle interior and windshield. They have used a Volterra Nonlinear Regressive with eXogenous inputs (VNRX) model where the inputs are the face features and the outputs are the point of gaze on the camera2 image.
% \cite{yuan2022self} proposed a self-calibrated gaze estimation system that eliminates manual calibration. It combines feature extraction from facial landmarks, head pose, and eye movements with unscented Kalman filtering for reliable tracking. Gaze pattern learning is then mapped using Gaussian Process Regression into different zones. 
\subsection{Simulated study on gaze object prediction by associating point of gaze on traffic object}
\par Gaze representation via gaze objects, some simulator-based studies regress the driver's gaze onto the scene (point of gaze), which further checks whether the PoG lies within any object's bounding box \cite{dua2020dgaze, deng2026cross}. In a lab setting, \cite{dua2020dgaze} developed a model to predict driver gaze from a projected video recorded during real-world driving, using the DGAZE dataset. 
After projecting this point onto the video, they check whether the predicted gaze point falls inside the bounding box of an object. If it does, the corresponding object class is detected. Similarly, Deng et al. \cite{deng2026cross} investigated gaze-based semantic object identification by determining the traffic object corresponding to a given gaze point in a road scene. Instead of collecting real driver gaze data, the study used BDD100K images and manually placed points on objects to simulate gaze coordinates. These coordinates were then provided to YOLOv13, SAM2-based methods, and Qwen2.5-VL models to identify the targeted gaze object.

\subsection{Recent development in transformer based gaze estimation in non driving applications}
\par In non-driving gaze-based applications, Transformer-based gaze estimation models have become increasingly popular in recent years, particularly for gaze object prediction, because they can effectively model the relationships between the driver's gaze-related features and multiple objects or regions in a scene. Unlike conventional point-of-gaze estimation, these models can predict which specific object the person is looking at, providing a more semantically meaningful representation of visual attention. In this literature, we consider several popular non-driving applications based on gaze-object prediction studies.
\cite{cheng2022gaze} developed a gaze estimation model using a pure transformer (GazeTR-Pure) and a hybrid transformer model (GazeTR-Hybrid). GazeTR-Pure includes only the transformer architecture for face feature extractions and gaze regression. While in GazeTR-Hybrid, features are extracted using CNN models, and the gaze is regressed using a transformer encoder block. \cite{li2025nonlinear} proposed a nonlinear multi-head cross-attention network with programmable gradient information for gaze estimation in terms of gaze vector. The first programmable gradient information feature extraction module was designed to extract multiscale gaze-relevant features from facial images. It consists of three blocks: the first captures large-scale facial contours and coarse textures, the second extracts medium-scale local and morphological features, and the third (FC) encodes global visual details, fine textures, and microexpressions. In a study \cite{li2026geometry}, a geometry-guided multimodal framework for loco pilot gaze target estimation was proposed, integrating RGB scene features, driver head position, and monocular depth information to model the spatial relationship between the gaze origin and the scene structure. The framework uses SimDINOv2 for visual feature extraction, multimodal geometry guidance, and cross-modal gated fusion to generate a 2D gaze heatmap, with the maximum response representing the predicted gaze point.
\par In parallel, transformer-based frameworks have also been explored for gaze object interactions \citep{nieva2025towards, li2026geometry, mathew2026gazevlm}. \cite{tonini2023object} proposed an object-oriented gaze target detection method using an end-to-end Gaze Transformer. The model has three main components: (a) an Object Detector Transformer to detect all objects, including heads, (b) a Gaze Cone Predictor that generates a gaze vector and corresponding cone for each detected head, and (c) the Gaze-Object Transformer (GOT), which models relationships between gaze cones and detected objects. Similarly, in a retail environment, \citep{wang2024transgop} proposed TransGOP, a transformer-based framework for predicting gaze object interactions. The model is composed of two complementary components: an object detection branch and a gaze regression branch. The object detector processes the full scene image to localize and classify objects using a transformer-based detection backbone. Meanwhile, the gaze regressor takes both the head and scene images as inputs to generate a gaze heatmap. 
\subsection{Summary}
\par Based on the above discussion, research on gaze object prediction remains limited, with only a few studies \citep{dua2020dgaze, deng2026cross} exploring this problem, primarily in simulated or controlled environments. These studies typically assume that the gaze point is already known and subsequently map the pre-determined gaze point onto objects in the scene image; therefore, they primarily perform gaze-point-to-object association rather than explicitly predicting the gaze object from visual cues. Moreover, their reliance on simulated or controlled laboratory settings limits their applicability to real-world driving environments. To the best of our knowledge, no existing study using real-world driving data explicitly incorporates traffic-object information to develop a gaze object prediction model. To address these limitations, the present study proposes a transformer-based framework that directly predicts the driver's gaze object by jointly leveraging driver facial cues and traffic-object information in real world driving environments.

%The remarkable performance of transformers in various application domains inspired us to investigate their use for predicting the driver gaze object. So in this study we have proposed a gaze estimation model which directly predict the driver gaze object on which driver is looking while driving.

\section{Dataset}
\label{dataset}

Deep learning-based gaze estimation models, particularly those relying on convolutional neural networks (CNNs) and vision transformers, require large-scale and diverse datasets to achieve robust performance on driver gaze estimation tasks. The data required to train these models should ideally be collected in real driving environments,  which naturally include practical challenges such as reflections from sunglasses, facial glare from sunlight, and poor illumination under low-light conditions. Such factors are typically absent or inadequately represented when data are collected in simulated driving environments. In this study, we created a real-driving gaze dataset consisting of driver face-scene image pairs, detected object bounding boxes in the scene images, gaze-object labels, and 2D gaze points relative to the forward-facing scene image. LBW is an existing similar benchmark dataset. However, the traffic density in LBW is substantially low, and also, the number of traffic objects is mostly dominated by cars and trucks. Our UD-FSG dataset has been collected in a high-density urban driving environment with heterogeneous traffic, which makes gaze-object prediction challenging.
In this section, we describe the UD-FSG dataset, including the sensor setup and synchronization, driver characteristics, data collection procedure, and the dataset composition. Each of these aspects is discussed in detail below.
%consisting of driver face–scene image pairs and corresponding 2D gaze coordinates. The dataset, called the UD-FSG (Urban Driving–Face Scene Gaze) dataset, is the first dataset collected in a heterogeneous traffic environment and the second publicly available dataset after the existing LBW (Look Both Ways) benchmark dataset. 
% In this section, we first describe the sensor setup and sensor synchronization, drivers characteristics, data collection procedure, and, finally, the dataset used in this study. Each of these aspects is discussed below in detail. 

%\subsection{Data Collection}

\subsection{UD-FSG dataset}
%\subsubsection{Route Survey}
% Before start the data collection, we mark a route on Google Map and then verified it on the actual ground. For that we have conducted a reconnaissance survey first and understand the traffic conditions including traffic jams, traffic density, traffic light running status etc. Then after we have modified accordingly our route map and conduct a preliminary survey to understand the traffic situation and approx total travel time for each driver. Finally, based on the survey analysis, we selected an approximately 40 km route within Kanpur city.
 
 %which includes approx 35 turns on intersections as shown in Fig. \ref{fig:route}. 

% \begin{figure}[!htbp]
%     \centering
%     \includegraphics[width=0.8\textwidth]{Figures/Route.png}
%      \caption{Selected route for data collection}
%     \label{fig:route}
% \end{figure}

\subsubsection{Sensors setup and synchronization}
\begin{figure}[!htb]
    \centering
    \includegraphics[width=0.75\textwidth]{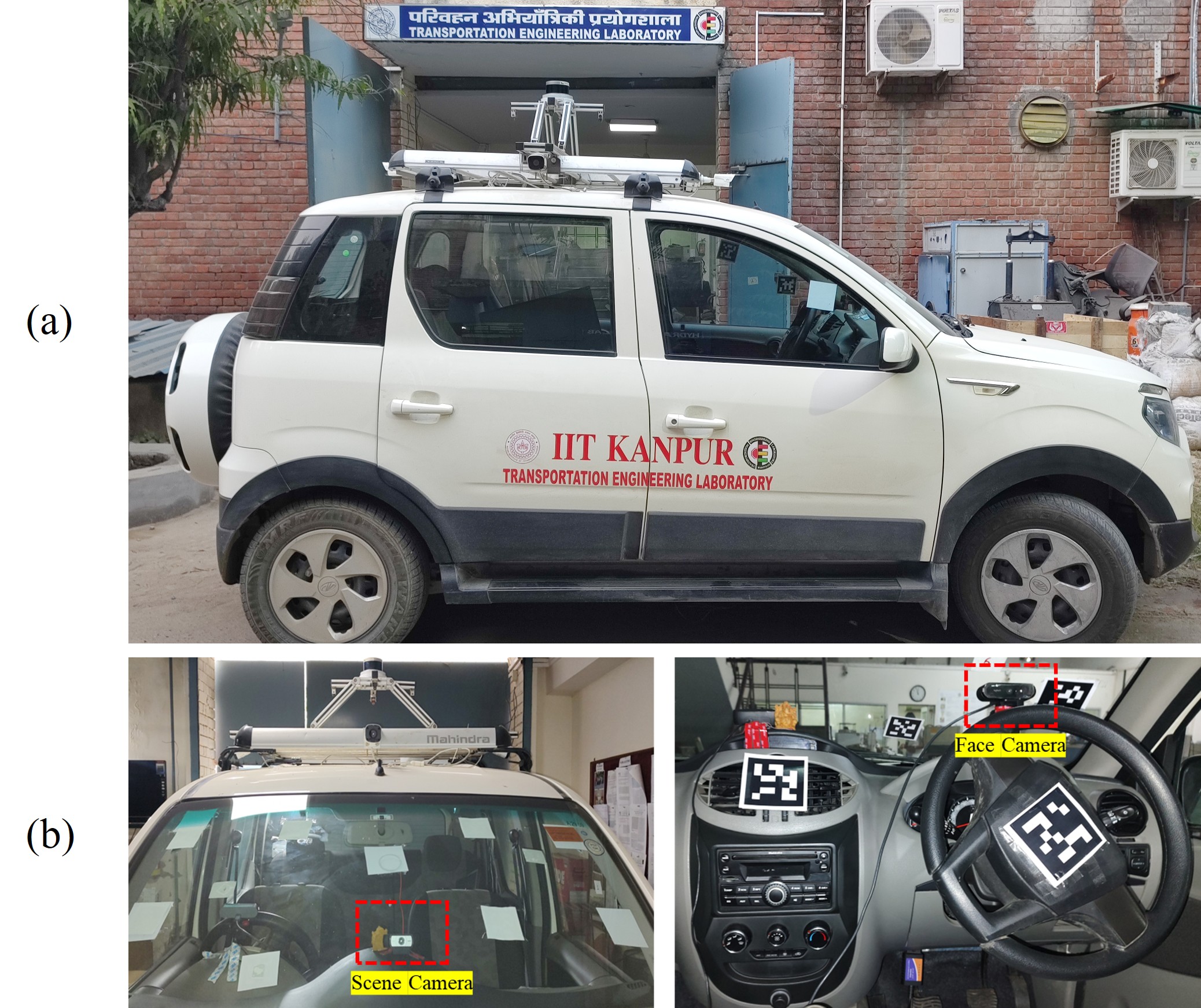}
     \caption{Data collection setup (a) Instrumented vehicle (b) Face camera and scene camera to capture driver face image and corresponding scene image}
    \label{fig:Sensor_Setup}
\end{figure}

The vehicle used for data collection in this study is an Instrumented Vehicle (IV), as shown in Figure~\ref{fig:Sensor_Setup} a. The Instrumented Vehicle is an SUV equipped with several sensors, including LiDAR (Light Detection and Ranging), cameras, OBD (On-Board Diagnostics), GPS (Global Positioning System), IMU (Inertial Measurement Unit), and Eye Tracker. LiDAR is used to measure distances and relative velocities of surrounding vehicles and pedestrians from our instrumented vehicle. At the same time, cameras capture video of the surrounding traffic environment and the driver's face. OBD records vehicle kinematics such as speed and acceleration, while GPS and IMU provide the vehicle's location, motion, and orientation, respectively. An eye tracker was used to capture the gaze information of the driver in terms of 2D gaze coordinates with respect to the eye tracker scene camera, which further transforms the 2D gaze coordinates with respect to the dashboard forward scene camera. Although data were collected from all of these sensors, this study utilizes only the face camera, the forward-facing scene camera as shown in Figure~\ref{fig:Sensor_Setup}b, and the pupil invisible eye tracker \citep{tonsen2020high} . The face camera records the driver's facial appearance, the scene camera captures the forward view, and the eye tracker provides ground-truth gaze coordinates for model development and evaluation.
All cameras were synchronized using a GStreamer application, while the cameras and eye tracker were synchronized using timestamps referenced to a stopwatch.

\subsubsection{Participants/Drivers}
The data was collected using professional drivers in real driving conditions in Kanpur city. We obtained permission from the Institute Ethics Committee to collect the driver's driving data before the study commenced. Before data collection, each driver provides their written informed consent.  
A total of 41 male professional drivers were recruited to participate in the study. However, due to a failure of one of the data storage devices, data from six participants were lost and could not be included in the analysis. Consequently, the final dataset consists of data from 35 participants. The participants had a mean age of 35.77 years (standard deviation (SD) = 6.30, range = 25–51 years) and a mean driving experience of 13.70 years (SD = 5.85, range = 3–30 years). It can be noted that the percentage of professional female drivers in the data collection region is extremely low \cite{tayal2025road}.
\subsubsection{Driver gaze ground truth creation}
\begin{figure}[!htbp]
    \centering
    \includegraphics[width=1.0\textwidth]{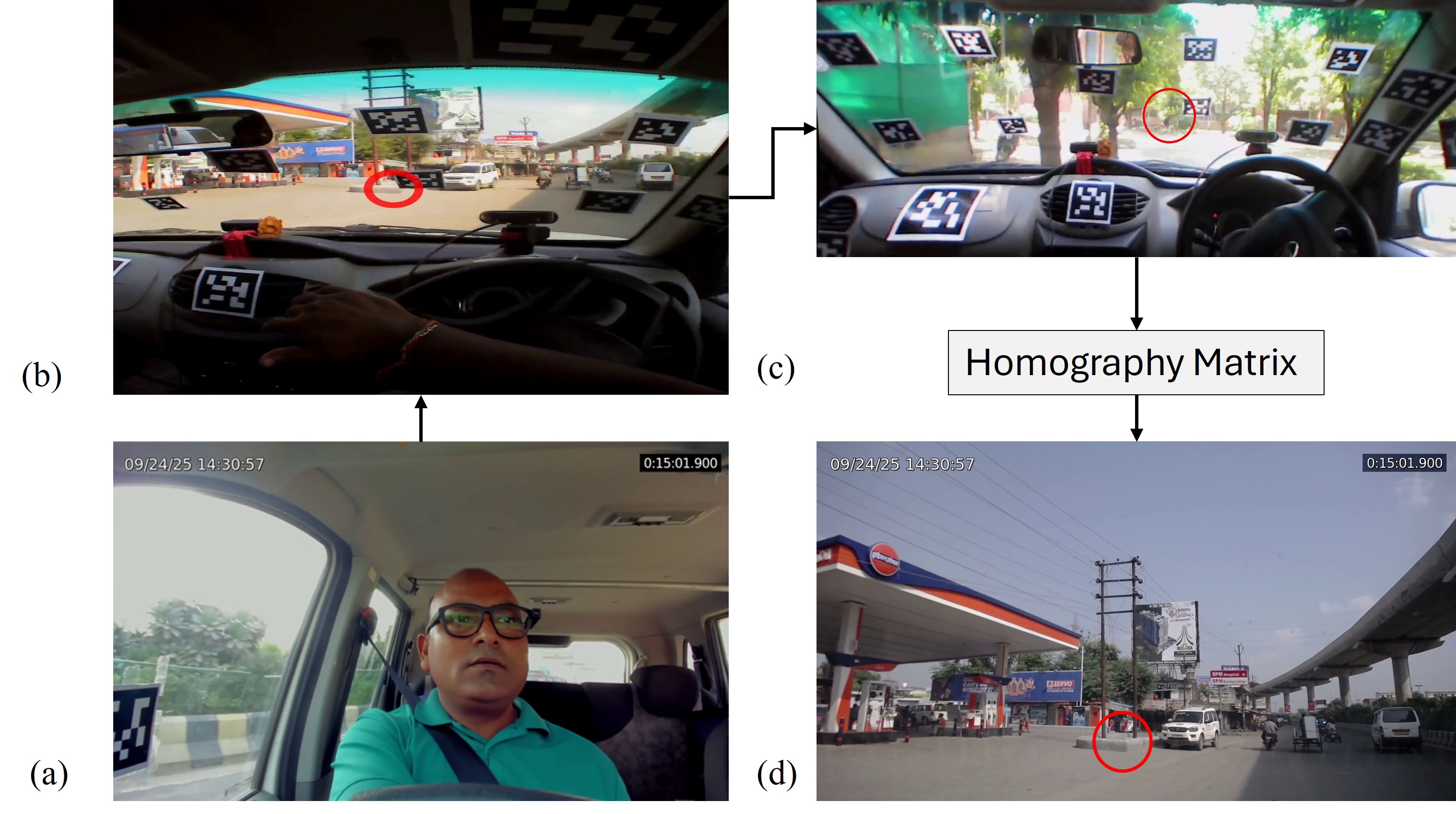}
     \caption{Illustration of transformation of gaze coordinate from eyetracker scene camera image to dashboard scene camera image}
    \label{fig:Gaze Transformation}
\end{figure}

The data were collected in an urban real-driving environment in Kanpur city, across arterial and sub-arterial roads at different times of the day to account for the effects of varying traffic density and sunlight conditions on the driver's face. Each driver drove the vehicle for approximately 1 hour, covering a travel distance of about 30–35 km. The video data of the driver's face and scene were originally recorded at 10 frames per second (fps), and frames were extracted at 5 fps to create this dataset. The extracted frames from the face and scene include the driver's face images, forward-view scene images. The gaze coordinates (2D gaze point with respect to eye tracker scene) of the driver corresponding face-scene image pairs were obtained from the eye tracker. Since the eye tracker scene camera is fixed to the eye tracker frame, which the driver wears, its position and orientation continuously change as the driver's head rotates to check the side wing mirrors, rear-view mirror, etc. Consequently, the gaze coordinates obtained from the eye tracker are expressed in a moving camera coordinate system as shown in Figure~\ref{fig:Gaze Transformation}a
\par To obtain gaze coordinates in a consistent reference frame, the gaze coordinates are transformed from the eye tracker scene camera coordinate system to the coordinate system of the fixed dashboard mounted scene camera, as shown in Figure~\ref{fig:Gaze Transformation}b. This transformation ensures that all gaze coordinates are represented relative to a fixed camera, enabling consistent analysis across drivers and driving sessions. The following procedure transformed the 2D gaze coordinate from the eyetracker scene camera to the dashboard scene camera.

\par
\begin{enumerate}[itemsep=0pt,parsep=0pt,topsep=0pt,partopsep=0pt]
\item Visual markers (AprilTags) were affixed to the windshield and side windows of the instrumented vehicle as shown in Figure~\ref{fig:Gaze Transformation}c. The positions of these AprilTags remained fixed throughout the entire data collection process, providing a consistent reference frame for gaze transformation.
\item Before each driving session, the windshield area of the instrumented vehicle was scanned using the eye tracker. During post-processing, a reference frame was selected from the scanned and recorded video of the forward windshield region of the instrumented capture by the eye tracker's scene camera. The reference image consists of several AprilTags pasted on the windshield of the car. The reference image is uploaded to the Pupil Cloud (a Pupil Lab cloud database for eye tracker gaze data analysis) visualization toolkit, and then the visualization toolkit transforms the 2D gaze coordinates recorded by the eye tracker (w.r.t. scene camera) from the eye tracker scene camera coordinate system to the fixed reference image, as illustrated in Figure~\ref{fig:Gaze Transformation}(b).
\item After obtaining the gaze coordinates in the fixed reference image as shown in Figure~\ref{fig:Gaze Transformation}c, these coordinates were further transformed to the dashboard forward scene camera coordinate system using a homography transformation defined by Equation~\eqref{eq:homography}. To compute the homography matrix, we selected approximately 10–12 image pairs for each participant, consisting of an eye-tracker-fixed reference image and a dashboard forward-scene camera image, with known gaze points in each pair. Please note that the reference image gaze coordinates were obtained from the Pupil Cloud (Pupil Labs) eye-tracker toolkit visualization discussed above. In the corresponding forward-scene image, at the exact location that matched the scene in the eye-tracker image, a circle was drawn using a photo multi-tool application, with the gaze coordinate as the center. The gaze point identified in the forward scene camera image was normalized with respect to the image dimensions as $x_f=x_f^{p}/W_f$ and $y_f=y_f^{p}/H_f$, where $x_f^{p}$ and $y_f^{p}$ denote the pixel coordinates of the gaze point, and $W_f$ and $H_f$ are the width and height of the forward-camera image, respectively. A $3\times3$ homography matrix, $\mathbf{H}$ was estimated using the corresponding eye tracker fixed reference image, and normalized forward scene camera image coordinates through RANSAC \citep{fischler1981random} based homography estimation.

% \begin{equation}\label{eq:homography}

% $H$=
% \begin{bmatrix}
% h_{11} & h_{12} & h_{13} \\
% h_{21} & h_{22} & h_{23} \\
% h_{31} & h_{32} & h_{33} \\
% \end{bmatrix}
% \end{equation}

\begin{equation}
\label{eq:homography}
\textbf{H} =
\begin{bmatrix}
h_{11} & h_{12} & h_{13} \\
h_{21} & h_{22} & h_{23} \\
h_{31} & h_{32} & h_{33}
\end{bmatrix}
\end{equation}

For an eye-tracker gaze coordinate $(x_e,y_e)$, the corresponding normalized gaze location in the dashboard forward-camera scene image is obtained as
\begin{equation}
\begin{bmatrix}
x'_f\\
y'_f\\
w
\end{bmatrix}
=
\mathbf{H}
\begin{bmatrix}
x_e\\
y_e\\
1
\end{bmatrix},
\end{equation}
where $\mathbf{H}$ represents the transformation from the eye-tracker coordinate system to the forward scene camera coordinate system. $w$ is the homogeneous scaling factor used to convert the transformed coordinates from homogeneous coordinates to Cartesian coordinates. The transformed coordinates are obtained by normalizing the first two components by $w$. 
\begin{equation}
x_f=\frac{x'_f}{w}, \qquad
y_f=\frac{y'_f}{w}.
\end{equation}
Finally, the normalized coordinates are converted into pixel coordinates of the forward-camera image as
\begin{equation}
X_f=x_fW_f, \qquad Y_f=y_fH_f.
\end{equation}
Thus, the calibrated homography transformation maps the eye-tracker gaze location to the corresponding pixel location $(X_f,Y_f)$ in the forward scene image, which is subsequently used to identify the traffic object at which the driver is gazing.
\end{enumerate}
%Please not after transformation the gaze coordinate, it was also mannually check by drawing a circle by taking a center coordinate as the center of circle and then stitched the image of eyetracker scene camera frame as shown Figure~\ref{fig:Gaze Transformation}a and Figure~\ref{fig:Gaze Transformation}b together and check it manually. The images where transform gazecoordinates are not matche in both images are discarded from the dataset.
\par Please note that after transformation of the gaze coordinate, it was also manually checked to ensure that the transformed gaze coordinate is in the same location as obtained in the eye tracker image.

\par
\begin{figure}[!htb]
   \centering
        \includegraphics[width=1.0\textwidth]{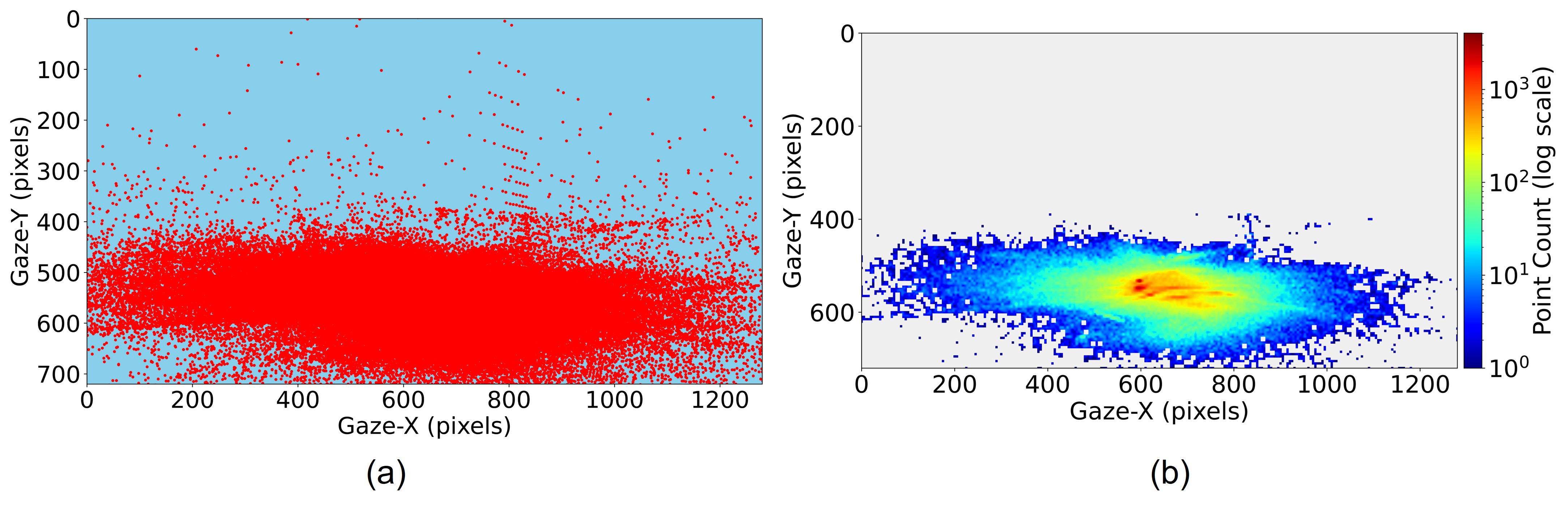}
        \caption{Visualization of driver gaze point distribution of UD-FSG dataset: (a) Gaze point locations on the scene across different drivers (b) Gaze density heatmap computed by aggregating point-of-gaze (PoG) coordinates over 5×5 pixel grid}
        \label{fig:data_char}
\end{figure}

 The distribution of driver gaze points in the scene is plotted in Figure \ref{fig:data_char}a by considering the image size equal to the scene image size  ($1280\times720$). To understand the spread of gaze points across the scene, a gaze density heatmap is also plotted by aggregating gaze points over $5\times5$ pixel region. Figure \ref{fig:data_char}b shows that driver gaze is concentrated in the middle regions (driver forward scene), which is expected, since the driver is predominantly looking in the forward regions while driving. However, our dataset also comprises images in which gaze is distributed across the lateral edges of the scene camera (i.e., $x < 200$ or $x > 1000$ pixels), making it suitable for modeling driver gaze toward the corners of the windshield. The UD-FSG dataset is available at the following link: \url{https://github.com/pavans20/Urban-Driving-Face-Scene-Gaze-Dataset.git}.

\subsubsection{Gaze object ground truth creation}
\label{Object_based_data_creation}
% We have trained a YOLO-v8 traffic Object detector. The details of the training is given in Appendix2. To detect the object from the scene image we have apply the data on the scene image detect the bounding box of object and count the number of object present in the scene image. Object detector was trained on different other dataset. We have detect the object form the each scene image. To make the ground truth of object on which driver is looking, the gaze coordinate we have map on the scene image and if the gaze point liying any object or the radius 20 pixels from the edges of the object than that object assigned as the gaze ground truth object. 
Gaze estimation via gaze-object prediction first requires detecting objects in a traffic scene. For this purpose, we trained a YOLOv8 (You Only Look Once version 8) based traffic object detector, the details of which are discussed in Section~\ref{Methodology}. The trained model has been applied to scene images to detect object bounding boxes and also determine the number of objects present in each scene.
Driver face image, scene image, and the corresponding ground truth gaze object are required to train the gaze-object prediction model. This ground truth gaze object can be a traffic object detected in a scene image or the background. Since an eyetracker provides gaze coordinates in the eyetracker scene camera images, which are then transformed into a dashboard scene camera image, as discussed above. The eye tracker does not provide direct gaze ground truth object information. So, to create a ground truth gaze object, the driver's gaze coordinates are mapped onto the traffic object detected in the scene image. If the mapped gaze point (gaze coordinate) on the scene image lies within a detected object's bounding box or 10 pixels apart from the nearest object boundary, that object is assigned as the ground truth gaze object. Otherwise, the ground truth is the background. For each scene image, we have the ground-truth gaze object index ID and the object's bounding box coordinates information. Please note that the background refers to the region of the scene image where no object bounding boxes are detected.
The number of face–scene frame pairs extracted at 5 FPS from the face and scene camera videos of all drivers was 373,488. Among these, 141,884 frame pairs (37.98\%) correspond to instances in which the driver's gaze falls on a detected traffic object. The remaining frames correspond to situations in which the driver's gaze is directed toward the background rather than a traffic object. 
\subsubsection{Dataset details}
\begin{figure}[!htb]
    \centering
    \includegraphics[width=1.0\textwidth]{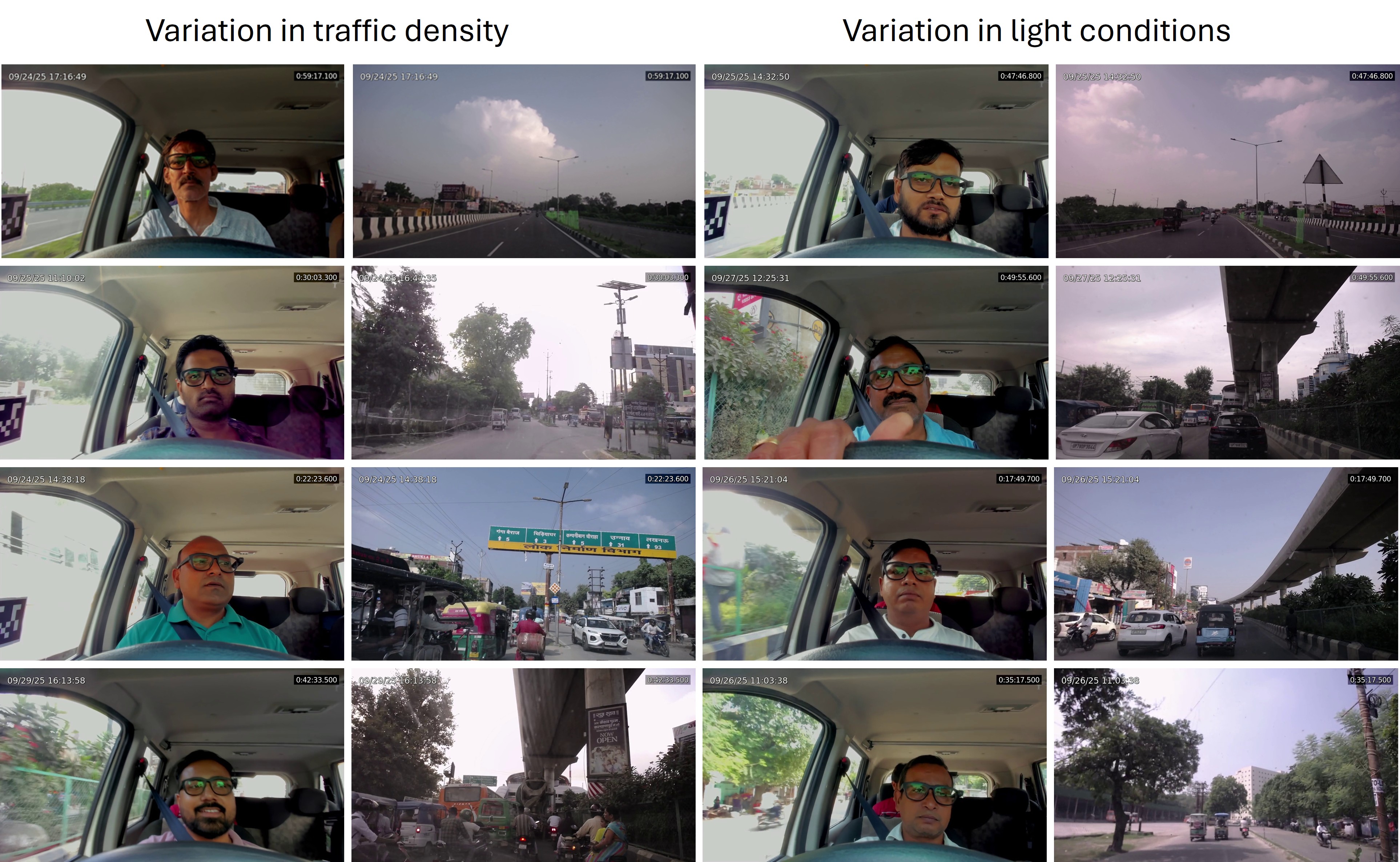}
        %\caption{Dataset showing different samples of drivers data with variations in traffic density and traffic light conditions.}
        \caption{Sample of the image pairs of face and scene showing variation in traffic density and the lighting conditions.}
        \label{fig:data_sample}
\end{figure}

%The data were collected in Kanpur city, India, across arterial and sub-arterial roads to incorporate variations in traffic density in the dataset. Data collection was conducted at different times of day to account for the effects of varying sunlight conditions on the driver's face. Each driver drove approximately one hour, covering a travel distance of about 30–35 km. The data were originally recorded at 10 frames per second (FPS), and frames are extracted at 5 FPS to create this dataset. 
% The proposed Urban Driving–Face Scene Gaze (UD-FSG) dataset is a benchmark real driving gaze dataset comprising data from 35 drivers, which includes 373,488 driver face and scene image pairs, and 2D gaze labels and first dataset collected in hetrogeneous traffic environment which consists of additionally detected traffic object bounding boxes in scene images, and gaze object labels. 
The proposed Urban Driving–Face Scene Gaze (UD-FSG) dataset is a benchmark real world driving gaze dataset comprising 373,488 driver face–scene image pairs from 35 drivers. 
A detailed comparison of our dataset with existing benchmark point of gaze datasets is presented in Table \ref{tab:gaze_dataset_comparison}. The existing driver gaze datasets are mostly for in-vehicle gaze estimation, comprising driver faces along with gaze ground truth as one of the regions of the vehicle interior, like the forward windshield, rear view mirror, etc. The details of these gaze datasets are provided in our previous study \citep{sharma2025evaluation}. To the authors' knowledge, the UD-FSG dataset is the second dataset consisting of synchronized driver face and traffic scene images, and 2D gaze coordinates ground truth labels, and first dataset collected in a heterogeneous traffic environment that additionally provides detected traffic object bounding boxes in scene images and gaze object labels. However, our dataset is significantly larger than one existing LBW dataset \cite{kasahara2022look} and includes more variation in traffic density and lighting conditions, as shown in Figure \ref{fig:data_sample}. The traffic environment consisting of diverse dynamic traffic agents (vehicles, pedestrians, etc.) makes the scene information meaningful and challenging enough to develop a robust driver gaze estimation model. Also, the eye tracker looks like regular prescription glasses, making the face image similar to that observed during real-world driving tasks. This data has been used to train our gaze object prediction model as discussed in Section \ref{Methodology}. 
% \par The distribution of driver gaze points in the scene is plotted in Figure \ref{fig:data_char}a by considering the image size equal to the scene image size  ($1280\times720$). To understand the spread of gaze points across the scene, a gaze density heatmap is also plotted by aggregating gaze points over $5\times5$ pixel region. Figure \ref{fig:data_char}b shows that driver gaze is concentrated in the middle regions (driver forward scene), which is expected, since the driver is predominantly looking in the forward regions while driving. However, our dataset also comprises images in which gaze is distributed across the lateral edges of the scene camera (i.e., $x < 200$ or $x > 1000$ pixels), making it suitable for modeling driver gaze toward the corners of the windshield. The UD-FSG dataset is available at the following link: \url{https://github.com/pavans20/Urban-Driving-Face-Scene-Gaze-Dataset.git}.

% bounding box information of the traffic objects, and corresponding gaze ground truth in terms of gaze object labels, along with 2D gaze points with respect to the forward scene image.

% \begin{figure}[!htb]
%    \centering
%         \includegraphics[width=1.0\textwidth]{Figures/Gaze_Point_Count.png}
%         \caption{Visualization of driver gaze point distribution of UD-FSG dataset: (a) Gaze point locations on the scene across different drivers (b) Gaze density heatmap computed by aggregating point-of-gaze (PoG) coordinates over 5×5 pixel grid}
%         \label{fig:data_char}
% \end{figure}

\begin{table}[!htbp]
\centering 
\begin{threeparttable}
\caption{Comparison of our dataset with existing benchmark driver gaze datasets.}
\label{tab:gaze_dataset_comparison}
\begin{tabular}{lcccccc}
\hline
Name & Face & Scene & Subjects & Size & Gaze GT \tnote{1} & Scenario \\ 
% \hline

% DrivFace \citep{diaz2016reduced} & Y\tnote{2} & N\tnote{3} & 4 & 606 & 3 & Real \\
% LISA GAZE v2 \citep{rangesh2020driver} & Y & N & 10 & 47k\tnote{4} & 7 & Parked \\
% DG-UNICAMP \citep{ribeiro2019driver} & Y & N & 45 & 1M\tnote{5} & 18 & Parked \\
% DGW \citep{ghosh2021speak2label} & Y & N & 338 & 50k & 9 & Parked \\
% DMD \citep{ortega2020dmd} & Y & N & 37 & 41h\tnote{6} & 9 & Real + Sim\tnote{7} \\
% ET-DGAZE \citep{sharma2025evaluation} & Y & N & 40 & 12k & 9 & Parked \\
\hline
DR(eye)VE \citep{palazzi2018predicting} & N \tnote{2} & Y \tnote{3} & 8 & 555k \tnote{4} & PoG & Real \\
LBW \citep{kasahara2022look} & Y & Y & 28 & 123k & PoG & Real \\
%\hline
\textbf{UD-FSG (Ours)} & Y & Y & 35 & 373k & PoG + Gaze Object & Real \\
\hline

\end{tabular}
\begin{tablenotes}[para,flushcenter]
\footnotesize
\setlength{\labelsep}{0.6em}
\setlength{\itemsep}{1pt}
%\item[1] Gaze Zone Count;
\item[1] Ground Truth
\item[2] No;
\item[3] Yes;
\item[4] Thousand;
%\item[5] Million;
%\item[6] Hour;
%\item[7] Simulated
\end{tablenotes}
\end{threeparttable}
\end{table}

\subsubsection{Dataset characteristics}
\label{Dataset_Characterstics}
\begin{figure}[!htb]
    \centering
    \includegraphics[width=0.80\textwidth]{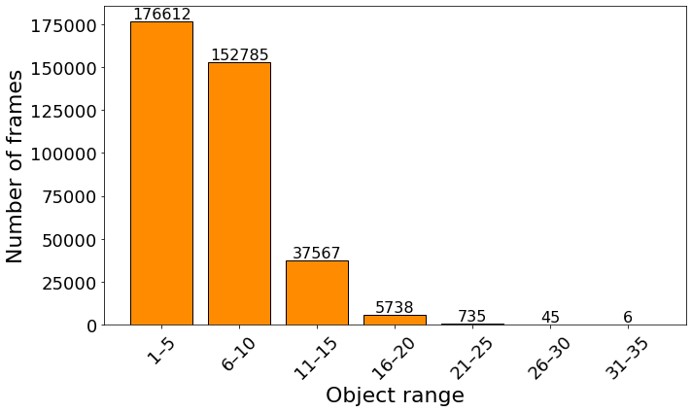}
        %\caption{Dataset showing different samples of drivers data with variations in traffic density and traffic light conditions.}
        \caption{Histogram present data object range in each scene frame}
        \label{fig:Object_histograme}
\end{figure}

\begin{figure}[!htb]
    \centering
    \includegraphics[width=1.0\textwidth]{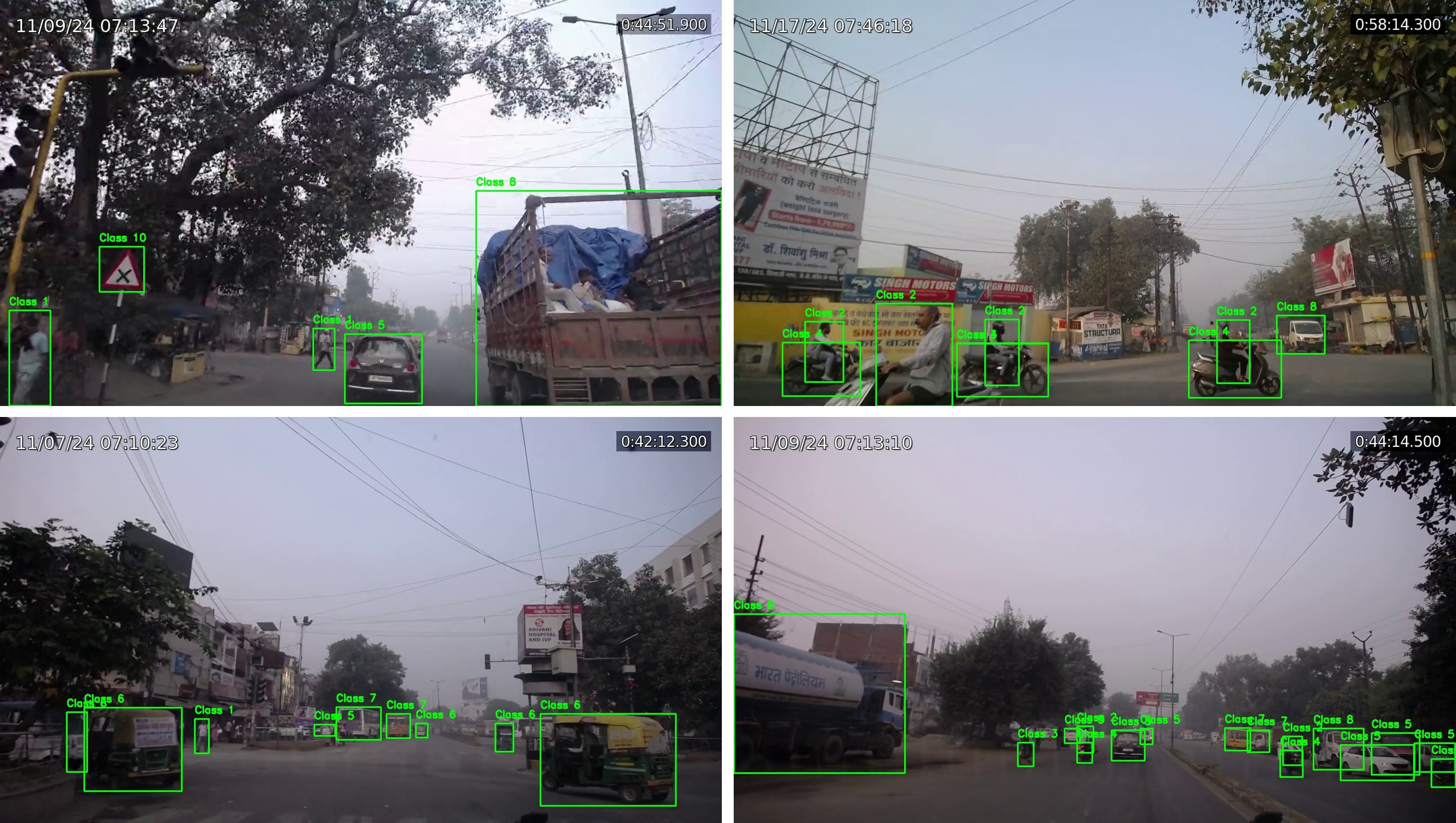}
        %\caption{Dataset showing different samples of drivers data with variations in traffic density and traffic light conditions.}
        \caption{Sample traffic scene images showing detected bounding boxes.}
        \label{fig:Object_detected_scene}
\end{figure}

Figure~\ref{fig:Object_histograme} illustrates the characteristics of the dataset in terms of the number of objects present in each scene. Out of a total of 373,488 frames, 176,612 frames contain objects between 1 and 5. Similarly, approximately 152,785 frames contain between 6 and 10 objects. The number of frames containing 16 to 20 objects is 5,738. Overall, the data show that approximately 99.78\% of the frames contain 20 or fewer traffic objects. The samples of detected objects shown in Figure~\ref{fig:Object_detected_scene} are arranged from the top-left to the bottom-right and contain an increasing number of detected objects.

\subsection{ Training dataset}% Benchmark DriveGaze-Object: A Driver Face–Scene Gaze Object Dataset}
% Finally, we created a benchmark dataset, DriveGaze-Object, a Driver Face–Scene Gaze Object Dataset comprising synchronized driver face images, forward scene images, detected object bounding box coordinates in the scene, and the corresponding ground truth gaze object. The gaze object refers to the traffic object or the background at which the driver is looking. The procedure of generating the gaze object annotations is described above. To the best of author's knowledge, DriveGaze-Object is the first benchmark gaze dataset that simultaneously provides driver face, scene and ground truth gaze objects. 

In the UD-FSG dataset, the number of background gaze labels is comparatively higher than the number of gaze-on-traffic-object labels, as discussed above. A subset of the background samples was selected to achieve a more balanced dataset for training our proposed TransGaze-Object model. To train this model, the dataset contains 189,850 synchronized face–scene image pairs, along with corresponding detected object bounding-box coordinates and ground truth gaze-object labels. The gaze object belongs to one of 11 classes: 10 traffic object categories (pedestrian, rider, bicycle, motorcycle, auto-rickshaw, car, bus, truck, traffic sign, and traffic light) and a background class. Among the 189,850 face–scene image pairs, 141,884 samples correspond to cases where the driver's gaze falls on one of the traffic objects, i.e., the ground truth is one of the traffic objects, while the remaining 47,966 samples correspond to the background class. Finally, we considered 26 drivers for the training set (165,969 samples), 5 drivers for the validation set (9,208 samples), and 3 drivers for the test set (14,673 samples). The details of the training dataset are given in the Table~\ref{tab:Train_Val_Test}. The next section details the methodology used to develop and train the gaze object prediction model.
%The next section discusses the detailed methodology used for the development and training of the gaze object prediction model.

\begin{table*}[ht]
\centering
\caption{Data used for training, validation, and testing}
\label{tab:Train_Val_Test}
\begin{doublespace}
\setlength{\tabcolsep}{24pt}
\begin{tabular}{lccc}
\hline
Ground Truth  & Train  & Val  & Test  \\ \hline
Traffic Object    & 125565 & 6441 & 9878  \\
Background        & 40404  & 2767 & 4795  \\ \hline
\textbf{Total}    & \textbf{165969} & \textbf{9208} & \textbf{14673} \\ \hline
\end{tabular}
\end{doublespace}
\end{table*}

\section{Methodology}
\label{Methodology}
% Driver gaze estimation in terms of point-of-gaze estimation, only reveals the gaze as a point in the scene image. Although, it provides a finer representation of gaze, it does not reveal the object the driver is looking at. Estimating gaze in terms of objects is essential for understanding driver visual attention, situational awareness, and the underlying decision-making process. To overcome the limitations of point-of-gaze estimation and also the important use of gaze object information in different driver safety applications motivates us to develop a model that can directly predict the object at which the driver is looking. 

\par In this section, we present a Transformer based Gaze Object prediction model (TransGaze-Object) that uses facial and spatial scene-object information to predict the object the driver is looking at. %The data used for training is discussed in the Section~\ref{dataset}.  We then discuss the performance of the proposed model, followed by a detailed analysis of failure cases.

\begin{figure}[!htbp]
    \centering
    \includegraphics[width=1.0\textwidth]{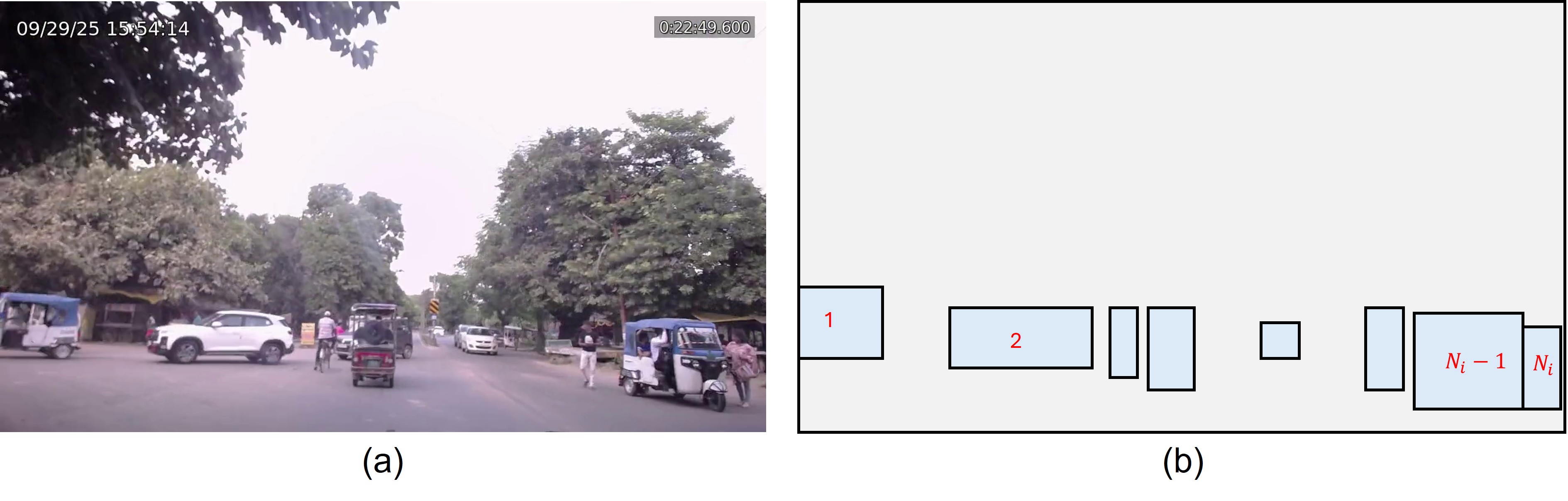}
     \caption{(a) Real image of scene containing the traffic objects and (b) Schematic representation of object present on the scene}
    \label{fig:Scene_Object_Representation}
\end{figure}

\textit{Problem formulation:}
%We propose a scene object-aware driver gaze estimation model in which the driver's gaze is represented by the object at which the driver is looking. 
Let us assume a given image $i$ as shown in Figure~\ref{fig:Scene_Object_Representation}a consists of $N_i$ traffic objects, as shown in Figure~\ref{fig:Scene_Object_Representation}b. The driver may look at any one of the $N_i$ traffic objects or background. Therefore, our objective is to estimate which of the $N_i+1$ traffic objects (+1 for background) the driver is looking at.

Let the face representation be defined as:
\begin{equation*}
F \in \mathbb{R}^{H \times W \times 3}
\end{equation*}
%where $T_f$ corresponds to the number of tokens (four face features extracted from different stages of feature extractor and two eye features) and $d$ is the embedding dimension.
% The scene is represented as a set of $N+1$ objects which includes maximum $N$ traffic object and 1 background. Since each scene image number of object presents are varying. Here $N$ is decided based on majority of scene image data in which maximum number of traffic objects can be $N$. 
where $H$ and $W$ are the height and width of the image, respectively, and 3 represents the RGB (Red-Green-Blue) channels. % $w$ width corresponds to the number of tokens (four face features extracted from different stages of feature extractor and two eye features) and $d$ is the embedding dimension.
The scene is represented as a set of $N+1$ objects, comprising a maximum of $N$ traffic objects and one background object. Since the number of traffic objects varies across scene images, a fixed value of $N$ is selected based on the maximum number of traffic objects present in the majority of the dataset. Scene images containing fewer than $N$ traffic objects are padded with virtual objects to maintain a consistent input representation. The objects are represented as:

% \begin{equation}
% B = \{b_i\}_{i=1}^{N+1}
% \end{equation}

\begin{equation*}
B = \{b_i\}_{i=1}^{N+1}, \quad b_i = (x_i, y_i, w_i, h_i)
\end{equation*}

Each bounding box $b_i=(x_i, y_i, w_i, h_i)$ corresponds to the $i$-th detected traffic object, where $(x_i, y_i)$ denotes the center coordinates of the bounding box, and $w_i$, $h_i$ represent its width and height, respectively.  And the background will be explained later in the discussion of object feature extraction.

The task is to predict the gaze object which represent the object index id ($\hat{y} $):
\begin{equation}
\hat{y}  \in \{1,\dots,N+1\}
\end{equation}

% \begin{equation}
% gaze_Object = f(\mathbf{F}, \mathbf{S})
% \end{equation}

%This formulation allows the model to directly map facial cues to scene objects using transformer based attention mechanism without explicitly estimating gaze direction.

\begin{figure}[!htbp]
    \centering
    \includegraphics[width=1.0\textwidth]{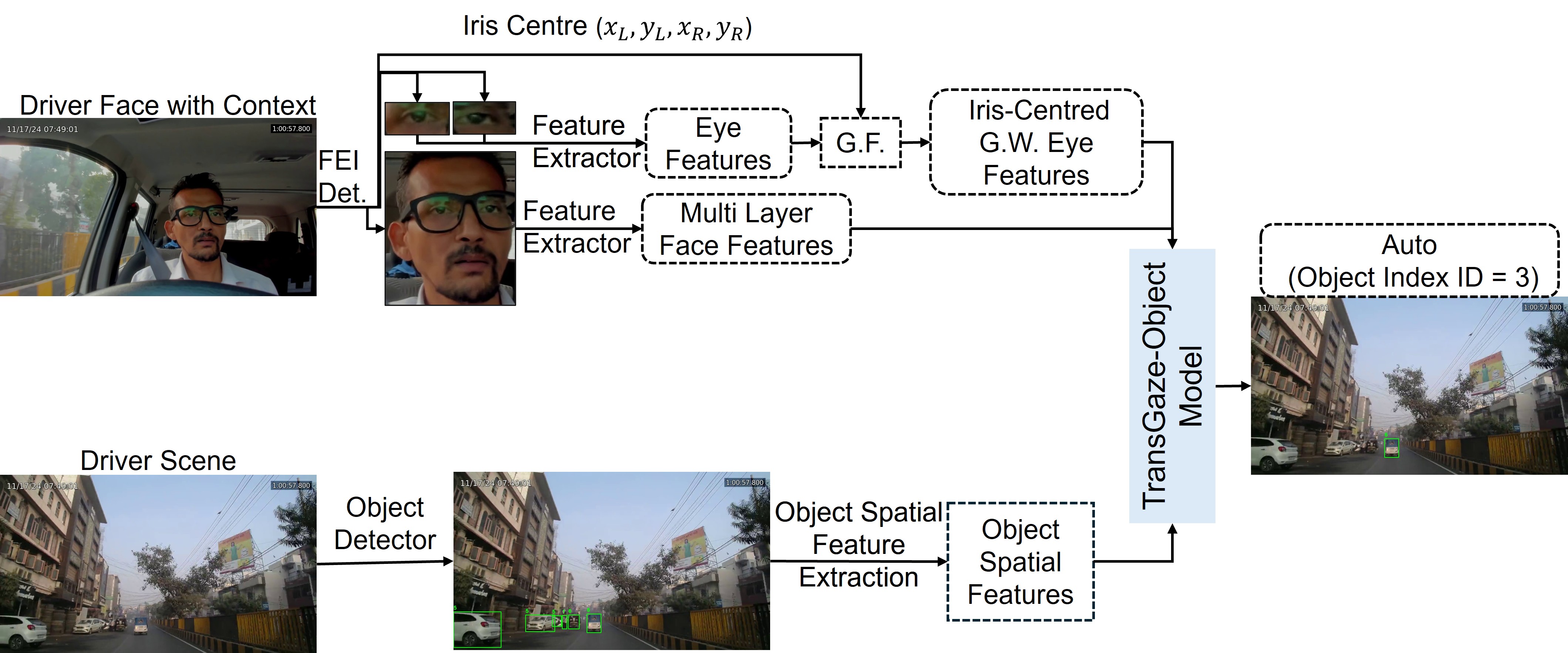}
     \caption{Illustration of overall gaze object prediction framework pipeline.}
    \label{fig:Gaze_Object_Prediction_Pipeline}
\end{figure}

% We propose a transformer based gaze object prediction model in which gaze is represented as the object the driver is looking at. That object is referred as the gaze object, and the corresponding gaze estimation task is referred as gaze object prediction. 
\textit{Overall framework:} The proposed model consists of five major components: (1) Facial geometry detection, including face, eye, and iris and scene object detection; (2) Feature extraction, including facial features extraction and object spatial and geometric feature extraction; (3) Transformer-based features encoding of face and scene object; (4) Cross attention between encoded facial and scene features; (5) Gaze object prediction head. Each of these steps is discussed next one by one. The pipeline of the proposed gaze object prediction is shown in Figure \ref{fig:Gaze_Object_Prediction_Pipeline}.

\subsection{Facial geometry and object detection module}
The first step of our proposed methodology is to detect the driver face from the face camera image and traffic objects from scene camera image as shown in Figure~\ref{fig:Face_EYE_IRIS_Object_Detection}.
The face image captured by dash-cam face camera, used to extract the driver face geometry using face-eye-iris (FEI) detector. Similarly the scene contains the object which is detected using a custom traffic object detector the details of which is given below.

% \subsubsection{Facial geometry detection}
% The facial geometry including face, eye, and iris are detected using the Face-Eye-Iris (FEI) detector similar as described in Chapter~\ref{Chapter6}, Section~\ref{Facial Geometry Extraction Module} and also shown in Figure~\ref{fig:/Face_Eye_Iris_Detection}a.  The details of object detection is given below and sample of detected object is shown in Figure~\ref{fig:Face_EYE_IRIS_Object_Detection}b

\subsubsection{Facial geometry detection}

\begin{figure}[!htbp]
    \centering
    \includegraphics[width=1.0\textwidth]{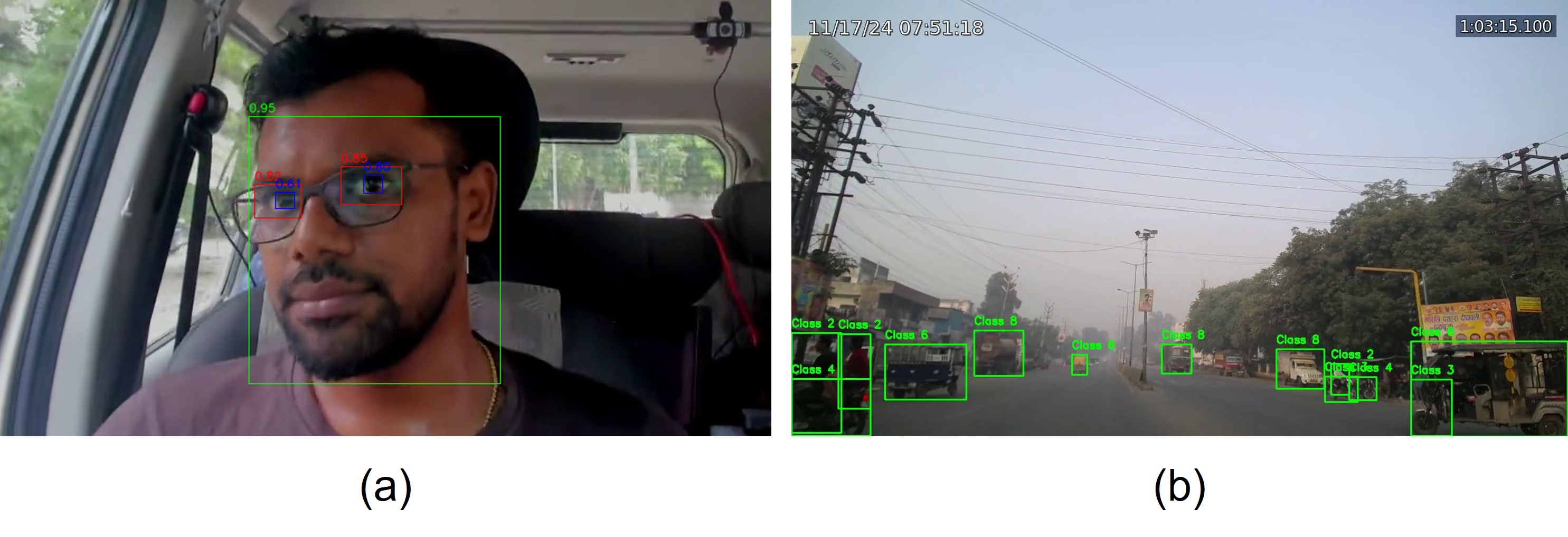}
     \caption{Schematic of detected (a) face, eyes, and irises and (b) objects in scene image}
    \label{fig:Face_EYE_IRIS_Object_Detection}
\end{figure}

The face camera captured the driver's face, which also includes some context of the surroundings, as shown in the Figure~\ref{fig:Face_EYE_IRIS_Object_Detection} a. To separate this context from the face and detect the face, eye, and iris, a custom face-eye-iris detector model was developed using a pretrained YOLOv8 model \citep{yolov8_ultralytics}. The model was trained using 481 drivers and 2200 annotated face images. The images of these drivers were taken from various existing driver gaze datasets such as DMD (Driver Monitoring Dataset) \citep{ortega2020dmd} DGAGE (Driver Gaze Mapping on Road) \citep{dua2020dgaze}, DGW (Driver Gaze in Wild) \citep{ghosh2021speak2label}, ET-DGAZE (Eye Tracker based Driver Gaze Dataset)\citep{sharma2025evaluation} and annotated using CVAT (Computer Vision Annotation Tool) into three classes: Face, Eye, and Iris. Note that the data used for the Face-Eye-Iris detector differs from the data used to train our gaze object prediction model.

The FEI model attained a mean Average Precision (mAP) of 95.7\% at an IoU threshold of 0.5, along with a recall of 93.0\%. This trained model is employed to localize the face, eyes, and iris regions from each input face image ($F$), as illustrated in Figure \ref{fig:Face_EYE_IRIS_Object_Detection}a. Based on the detected eye and iris bounding boxes, the iris center for each eye is calculated with respect to the upper-left corner of the corresponding eye region. When the iris is detected in only one eye, the missing iris location is estimated using the physiological property of conjugate eye movement \citep{yarbus2013eye, yang2019dual}, which assumes coordinated movement of both eyes. Consequently, the facial geometry extraction module provides the face bounding box, the left and right eye bounding boxes, and the corresponding left and right iris center coordinates expressed relative to their respective eye regions.

\par Under real-world driving conditions, there are situations in which the iris of both eyes cannot be reliably detected due to factors such as occlusion, motion blur, large head rotations, or varying illumination. To address these cases, a validity-aware gating strategy is incorporated into the framework. If valid iris coordinates are not available, the corresponding iris representation is deactivated, and its influence during feature fusion is eliminated to avoid introducing unreliable information into the gaze estimation process. The model then dynamically places greater emphasis on the remaining visual cues, including facial appearance, eye-region features, and scene information. This adaptive mechanism enables the framework to maintain stable, reliable gaze-estimation performance even when iris location information is unavailable.

% =========================================================

\subsubsection{Traffic object detection}
The proposed gaze object detection model requires the information of the traffic objects present in the traffic scene, captured by dashboard forward scene camera, as shown in Figure~\ref{fig:Face_EYE_IRIS_Object_Detection}b. Therefore traffic object detection is a prerequisite for the proposed model. We trained a YOLOv8 \citep{yolov8_ultralytics} object detector to detect traffic objects in scene images.
The traffic objects are categorized into 10 classes, namely: pedestrian, rider, bicycle, motorcycle, car, auto-rickshaw, bus, truck, traffic light, and traffic sign.

To train the traffic object detector, traffic data was obtained from open-source datasets, including the IDD (Indian Driving Dataset) \citep{varma2019idd}, the nuImages \citep{Caesar_2020_CVPR} dataset, UD-FSG dataset\citep{sharma2026sgap}. Since the original datasets do not contain annotations for all the desired traffic object categories (like nuImages does not contains rider, auto-rickshaw, traffic light, traffic sign class),  additional manual annotations were done using the CVAT (Computer Vision Annotation Tool) application. Specifically, two annotators labeled the missing classes to ensure consistent, uniform annotations across the entire dataset, making it suitable for training an object detection model. A pre-trained YOLOv8 model was fine-tuned and trained on 10,981 traffic-scene images containing multiple objects. The model achieved a mean Average Precision (mAP) of 79.1\% at a confidence threshold of 0.5 (50\%). Since object detection is not 100\% accurate, cases in which objects are not detected are manually annotated to prepare the data for the Gaze Object Prediction model. 

%Since the detection is not 100\% in such cases where detection is not occurred, manual annotator annotates those cases to make the data ready for Gaze Object Prediction model. 

\subsection{Feature extraction module}
The multi stream feature extraction module is designed to process the heterogeneous input modalities (i.e., different in their visual characteristics and semantic information) through separate feature extraction streams \citep{liu2025beyond}.  Each stream learns modality specific representations from the face, eye, and scene inputs. 

%\subsubsection{Face features and Gaussian-weighted eye features extraction}

\subsubsection{Face features extraction}

% \subsubsection{Face feature extraction}
% \label{Face Feature Extraction}
We used a pretrained ResNet-18 backbone as a hierarchical feature encoder to obtain compact and discriminative facial representations for gaze estimation. The detected face image is first resized to \(
I_f \in \mathbb{R}^{3 \times 224 \times 224}\) to fit the ResNet (Residual Network) architecture input configuration. This resized image is then  normalized and forwarded through the convolutional stem and the four residual stages of ResNet-18 \citep{he2016deep} are used to extract intermediate feature maps.
Here, the convolutional stem refers to the initial layers that extract low-level visual features, while the residual stages consist of stacked residual blocks (also called Layer-1/2/3/4) that progressively learn higher-level representations. This can be represented as:

\begin{equation}
F_l = \mathcal{B}_l(I_f), \quad l \in \{1,2,3,4\}
\end{equation}

where, \( \mathcal{B}_l(\cdot) \) denotes the transformation up to the $l^{th}$ residual block,  
\(
F_l \in \mathbb{R}^{C_l \times H_l \times W_l}
\) denotes each feature map
with channel dimensions 
\(
C_l \in \{64,128,256,512\} 
\) for \(
l \in \{1, 2, 3, 4\} 
\), and $H_l$, $W_l$ represent the spatial resolution (height and width) of the feature map.

Since the channel dimensions $C_l$ differ across layers, we project each feature map into a unified 256-dimensional embedding space using a learnable $1 \times 1$ convolution:

\begin{equation}
\hat{F}_l = \phi_l(F_l), \quad 
\hat{F}_l \in \mathbb{R}^{256 \times H_l \times W_l}
\end{equation}

where, $\phi_l(\cdot)$ represents the channel projection operation.

Finally, to obtain a compact global representation, adaptive global average pooling (GAP) is applied over the spatial dimensions to produce a 256-dimensional feature vector from each layer, represented as $f_l = \mathrm{GAP}(\hat{F}_l) \in \mathbb{R}^{256}$. Where each channel response is computed as:

\begin{equation}
f_l =
\text{GAP}(\hat{F}_l)_c =
\frac{1}{H_l W_l}
\sum_{i=1}^{H_l}
\sum_{j=1}^{W_l}
\hat{F}_l(c,i,j)
\label{eq:gap_feature}
\end{equation}

where, $c \in \{1, \dots, 256\}$ denotes the channel index of the projected feature map.

\begin{figure}[!htbp]
    \centering
    \includegraphics[width=0.85\textwidth] {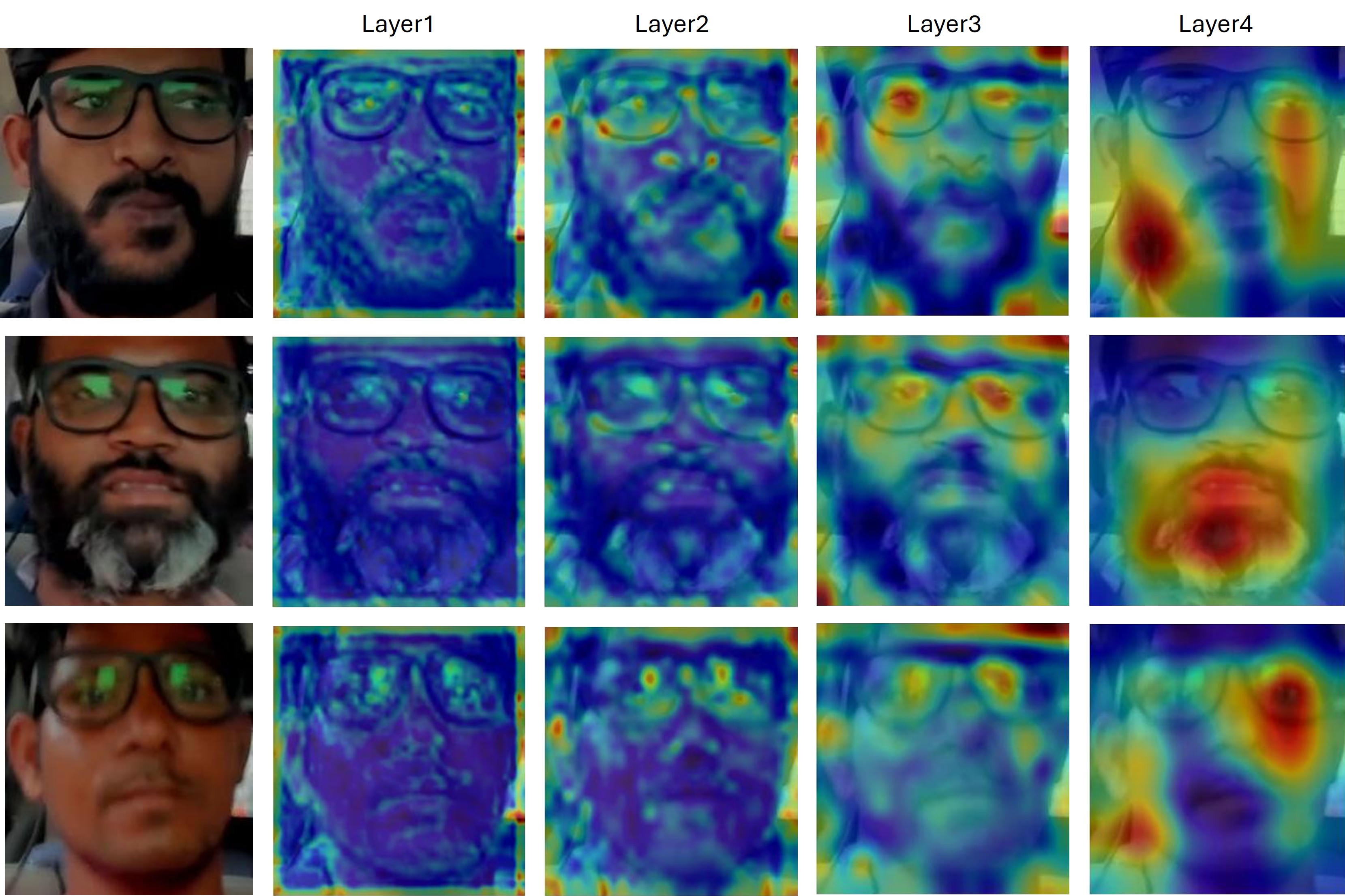}
     \caption{Visualization of face features extracted from different layers of ResNet-18 for three drivers.}
    \label{fig:Different_layer_features}
\end{figure}

Thus, for each face image, four hierarchical global facial feature vectors 
\(
\{f_1, f_2, f_3, f_4\}
\),
each of dimension 256, is extracted from different semantic depths of the network. These multi-level global embeddings capture complementary facial information, ranging from fine-grained texture patterns in shallow layers to high-level structural semantics in deeper layers, as shown in Figure \ref{fig:Different_layer_features}. The resulting 256-dimensional global representations are subsequently utilized in the proposed multi-modal feature fusion module for gaze object prediction.

\subsubsection{Gaussian weighted eye feature extraction}
\label{Gaussian-Weighted Eye Feature Extraction}
\begin{figure}[!htbp]
    \centering
    \includegraphics[width=1.0\textwidth] {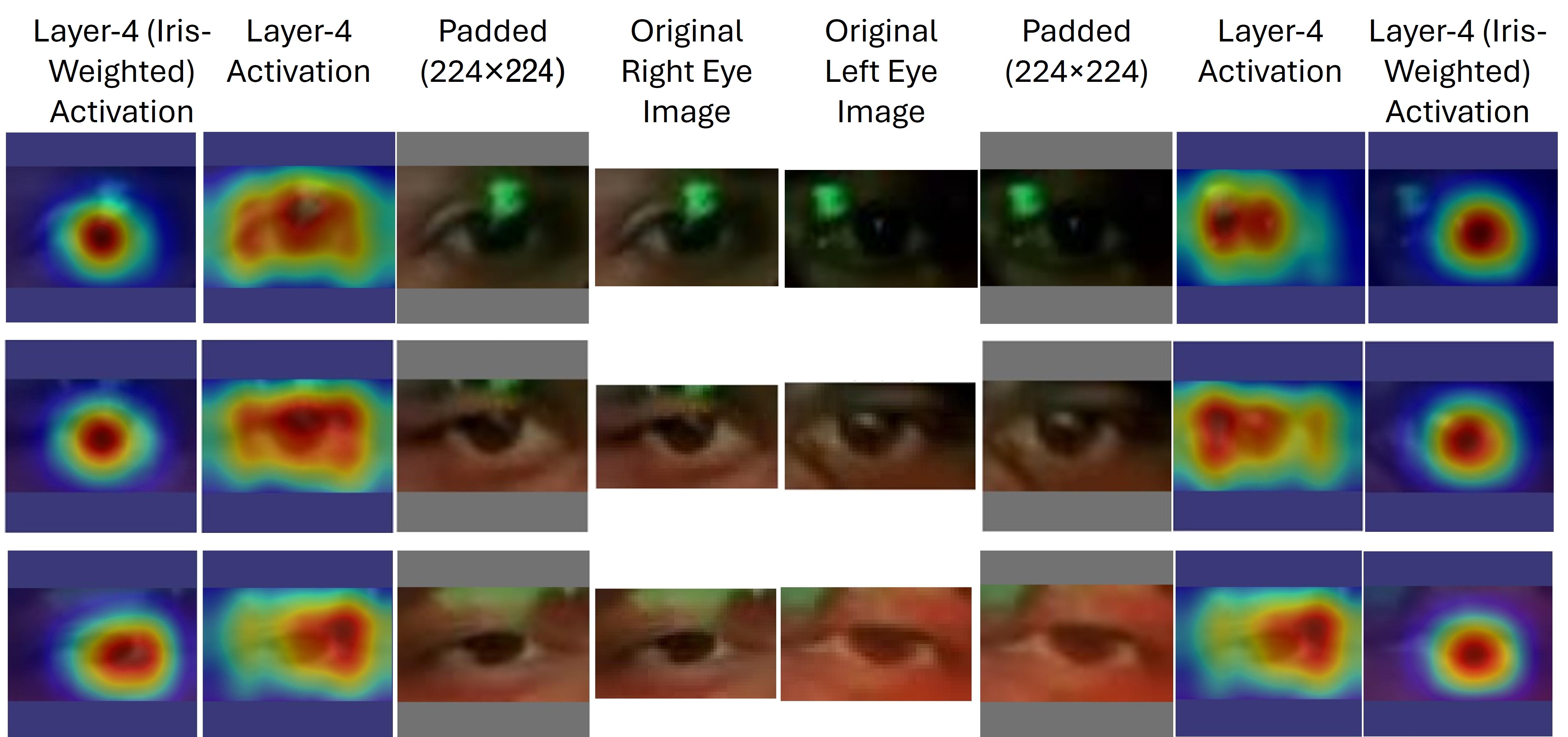}
     \caption{Visualization of eye features extracted from Layer-4 of ResNet-18 with constant padding value of 114, comparing unweighted and Gaussian-weighted feature responses emphasizing the iris region.}
    \label{fig:Eye Features of L4-Layer}
\end{figure}

Along with the overall face features, the eyes and the corresponding iris position with respect to the eye, are extremely important to determine the gaze location of the participants. Therefore, we design an efficient feature extraction of the eye region, detected using our FEI model, along with the iris position. First, the eye features are extracted using a pretrained ResNet-18 backbone, where the final residual block (layer4) is employed to obtain high-level semantic features. However, since the cropped eye images from FEI detection do not meet the required 224×224 input resolution of the ResNet architecture, we applied resizing (224×224) followed by constant padding to preserve the aspect-ratio requirement. A padding value of 114 is selected as a neutral gray intensity to minimize artificial boundary effects and prevent unintended feature activations in the padded regions. Figure \ref{fig:Eye Features of L4-Layer} shows a sample images of the original left and right eye images, along with the padded images used as input for ResNet model. The padded left and right eye images, with dimensions of each
\(
I_e \in \mathbb{R}^{3 \times 224 \times 224},
\)
is then normalized and forwarded through the convolutional stem and residual layers of ResNet-18 separately for both left and right eyes and producing a deep feature map for both left and right eye.% concatenating the left and right eye features:

\begin{equation}
F_e = \mathcal{B}_4(I_e), \quad 
F_e \in \mathbb{R}^{512 \times H_4 \times W_4}
\end{equation}

Similar to the facial feature extraction module, a $1 \times 1$ convolutional projection is applied to reduce the channel dimensions to 256.

\begin{equation}
\hat{F}_e = \phi_4(F_e), \quad 
\hat{F}_e \in \mathbb{R}^{256 \times H_4 \times W_4}
\end{equation}

To emphasize the iris region derived from FEI model within the eye feature map, a spatial Gaussian weighting function centered at the iris location is applied. Let $(c_x, c_y)$ denote the projected iris center coordinates in feature map space. The 2D Gaussian weight at spatial location $(x,y)$ is defined as:

\begin{equation}
G(x,y) = 
\frac{
\exp\left(
-\frac{(x - c_x)^2 + (y - c_y)^2}{2\sigma^2}
\right)
}{
\sum_{i=1}^{H} \sum_{j=1}^{W}
\exp\left(
-\frac{(i - c_x)^2 + (j - c_y)^2}{2\sigma^2}
\right)
}
\end{equation}

where, $\sigma$ (= 1.2 in our experiment) controls the spread of the Gaussian distribution. This normalized weighting map assigns higher importance to features closer to the iris center while suppressing peripheral regions.

The Gaussian-weighted feature map is then computed as:

\begin{equation}
\tilde{F}_e(c,x,y) = \hat{F}_e(c,x,y) \cdot G(x,y)
\end{equation}

where, $c$ denotes the channel index. Finally, a global representation is obtained via adaptive global average pooling: 

\begin{equation}
    E = \mathrm{GAP}(\tilde{F}_e) \in \mathbb{R}^{256}
    \label{eq:embedding_gap}
\end{equation}
where, $\mathrm{GAP}(\tilde{F}_e)$ is computed using similar equation \ref{eq:gap_feature}. Note that $E$ is represented for separate left and right eyes and indicated as $e_L$ and $e_R$. Figure \ref{fig:Eye Features of L4-Layer} provides the samples images for each step of eye feature extraction. This Gaussian-weighted global embedding enhances iris centered discriminative information while retaining contextual eye features, making it particularly suitable for precise gaze estimation.

The face features and eye features which are used to cues of gaze direction and used to find the similarity with the scene object present in scene image. The extracted features from face are represented as \(f_1, f_2, f_3, f_4 \in \mathbb{R}^{256}\) and the features from eyes are represented as \(e_L, e_R\in \mathbb{R}^{256}\). The four face features extracted from different stage or layers of ResNet-18 represents the local and global features of facial geometry and two eye features, extracted from layer 4 of of ResNet-18 represents the global features of the eye. Combining the face and eye features, the facial features set is defined as:  
\begin{equation}
F = \{f_1, f_2, f_3, f_4, e_l, e_r\} \in \mathbb{R}^{6 \times d}
\label{Eq:Facial_Features}
\end{equation}

So finally 6 number of tokens ($T_f = 6$) represented the facial features (four face features extracted from different stages or layer of ResNet-18 and represented the local and global features of the facial geometry and two eye features extracted stage 4 or layer 4 represented the global features of the eye).

\subsubsection{Scene object spatial features}
As discussed in Section~\ref{Dataset_Characterstics}, the dataset analysis shows that approximately 99.78\% of the frames contain 20 or fewer objects (traffic objects and background), as illustrated in Figure~\ref{fig:Object_histograme}. Therefore, each scene is represented as a fixed set of $N+1 = 20$ objects, comprising a maximum of 19 traffic objects and 1 background object. Images with traffic objects $> N= 19$ contains typically very small objects (because objects are very far from driver). Therefore for images, where number of objects $> 19$, the largest 19 objects in terms of area are considered. 
\par
\begin{figure}[!htbp]
    \centering
    \includegraphics[width=0.85\textwidth]{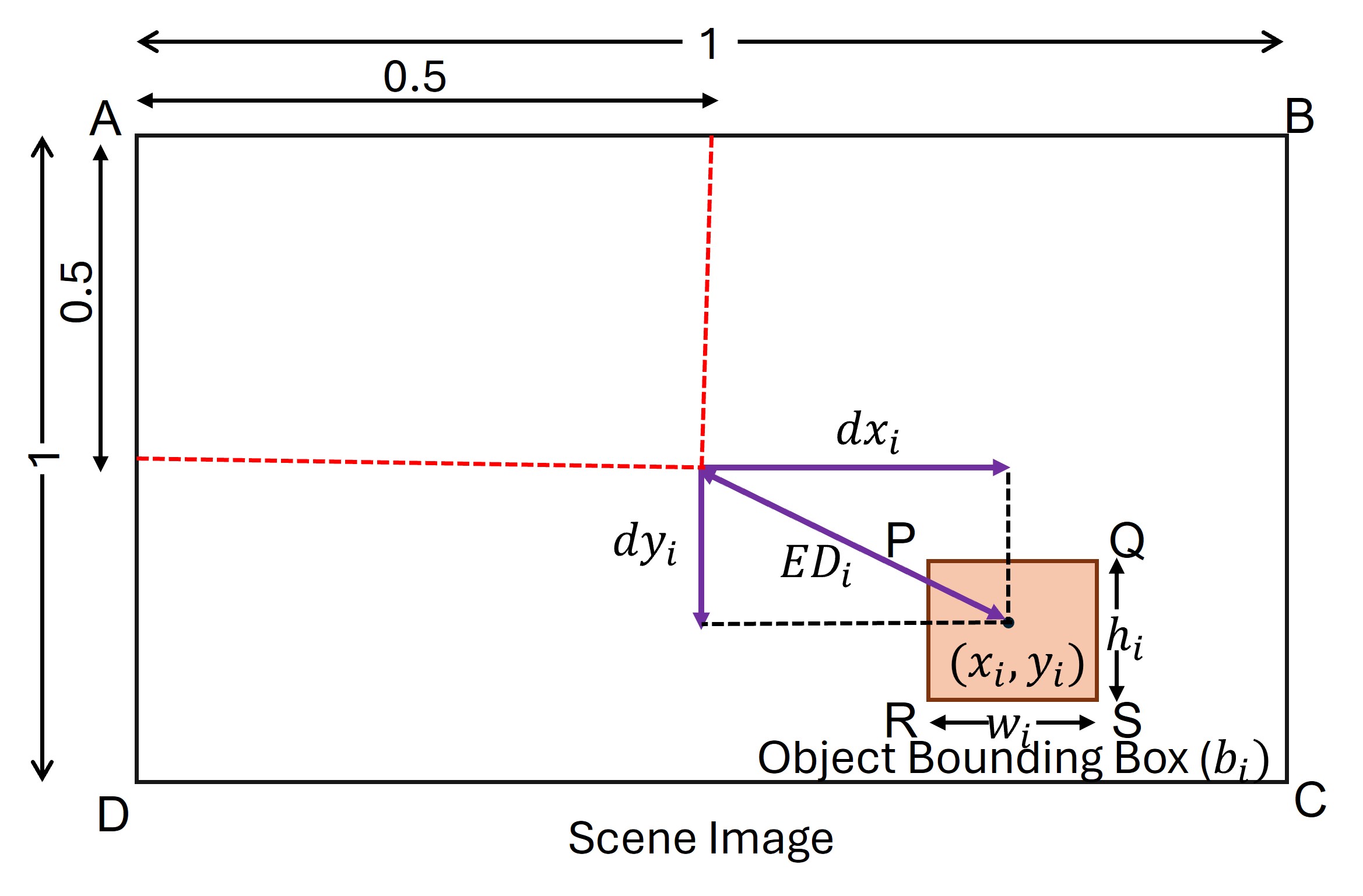}
     \caption{ A schematic representation to show the object features extracted}    \label{fig:Scene_Object}
\end{figure}

 Figure~\ref{fig:Scene_Object} illustrates the schematic representation of the spatial features extracted for a detected scene object. The rectangle $ABCD$ represents the complete scene image, while the orange rectangle PQRS denotes the object bounding box $b_i$. The point $(x_i, y_i)$ denotes the normalized center coordinates of the object bounding box, where $x_i$ and $y_i$ represent the horizontal and vertical center coordinates respectively. The normalized width and height of the bounding box are denoted by $w_i$ and $h_i$ respectively. The primary object features include the position of bounding box of the objects. This represented as:
\begin{equation}
B = \{b_i\}_{i=1}^{N+1}, \quad b_i = (x_i, y_i, w_i, h_i)
\end{equation}

To enhance object representation, each bounding box is augmented with additional spatial features. First area of each object is computed as: $a_i = w_i \cdot h_i$. To capture positional bias, offsets, and  Euclidean distance from the center of the scene image (0.5, 0.5) computed using $dx_i = x_i - 0.5, \quad dy_i = y_i - 0.5$, and $ED_i = \sqrt{dx_i^2 + dy_i^2}$ respectively. The demonstration of how the $dx_i$, $dy_i$ and $ED_i$ computed are shown in Figure~\ref{fig:Scene_Object}. 
Accordingly, the spatial features representation of the $i^{\mathrm{th}}$ object is expressed as:
% \begin{equation}
% s_i = [x_i, y_i, w_i, h_i, a_i, dx_i, dy_i, ED_i]
% \end{equation}

\begin{equation}
f_i^{obj} = [x_i, y_i, w_i, h_i, a_i, dx_i, dy_i, ED_i] \in \mathbb{R}^8
\end{equation}
This representation encodes not only the size of the object but also its relative position with respect to the image center, which is crucial since human gaze often exhibits center bias i.e driver is looking predominantly in forward and scene camera is placed at center of dashboard.

The spatial features are projected into a higher dimensional embedding space using a linear transformation:

\begin{equation}
s_i^{obj} = f_i^{obj} W_s + b_s
\end{equation}

where $s_i^{obj} \in \mathbb{R}^8$ represents the spatial feature vector, $W_s \in \mathbb{R}^{8 \times d}$ and $b_s$ is a learnable weight matrix and bias term.

A fixed spatial representation, \([-1,-1,2,2]\), was assigned to the background to distinguish it from valid traffic-object bounding boxes. These out-of-range coordinates and dimensions provide a unique spatial representation of the background while keeping it consistent across all scene images.

Please note that, to distinguish the background from valid traffic-object bounding boxes, a fixed spatial representation ([-1, -1, 2, 2]) was assigned to the background. The out-of-range coordinates and dimensions provide a unique spatial representation for the background. The remaining spatial features of the background were computed in the same manner as those of the traffic objects. Consequently, the final spatial feature vector for the background is ([-1, -1, 2, 2, 4, -1.5, -1.5, 2.12]), which remains fixed and consistent across all scene images.
% Please note that to distinguish the background from the valid traffic object bounding boxes a fixed spatial representation center coordinates \([-1, -1, 2, 2]\) was assigned to the background. The remaining spatial features of the background are computed in the same manner as for traffic objects, and this is fixed across all scene images. So the final features of the background become this [-1, -1, 2, 2, 4, -1.5, -1.5, 2.12] and consistent across all scene images.

Finally, the scene is represented by combining features from all objects (N+1 = 20), represented as: 
\begin{equation}
S^{obj}= \{s_1^{obj}, s_2^{obj}, \dots, s_{N+1}^{obj}\} \in \mathbb{R}^{{N+1} \times d}
\label{Eq:Scene_Features}
\end{equation}

This projection serves two key purposes. First, it aligns the spatial features with the embedding dimension of facial features, enabling meaningful interaction through attention mechanisms. Second, it allows the model to learn task-specific combinations of spatial attributes such as object position, size, and distance.

% =========================================================
%\subsection{Transformer-based Feature Encoding}
\subsection{Transformer-based encoding of face and scene features}
The proposed Transformer based Gaze Object (TransGaze-Object) prediction model is based on the hypothesis that accurate driver gaze object prediction requires jointly modeling the driver's facial cues and the surrounding scene objects. 
Multi-level facial features are extracted from the face and iris-weighted eye regions using separate pretrained ResNet-18 networks, while the spatial and geometric features of the detected scene objects are represented independently. The scene-object spatial and geometric features are subsequently transformed into the embedding dimension ($d$) using multi-layer perceptron (MLP), ensuring dimensional compatibility with the facial feature representations for subsequent attention-based fusion.
Since these features are extracted separately, they do not inherently capture the relationships within their respective feature sets. Therefore, two separate transformer encoder blocks (facial features and scene features) with self-attention are used to learn contextual dependencies within facial features and within scene object features. This enables the model to generate enriched feature representations by capturing interactions within each set of features.
\par After obtaining contextual facial and scene-object representations, the proposed model employs a cross-attention mechanism to establish relationships between the driver's gaze cues and surrounding scene objects. In this framework, the encoded facial features are used as the queries, while the encoded scene object features serve as the keys. This design is motivated by the fact that facial features encode the driver's visual intention, whereas scene objects represent potential gaze targets. The query–key matching mechanism enables the model to identify the object whose representation best aligns with the driver's gaze characteristics, and the resulting attention scores are used to predict the most likely gaze object. This cross-attention formulation constitutes the core contribution of the proposed TransGaze-Object model, effectively integrating driver appearance and scene context to predict gaze objects robustly.

\par
\begin{figure}[!htb]
    \centering
    \includegraphics[width=0.55\textwidth]{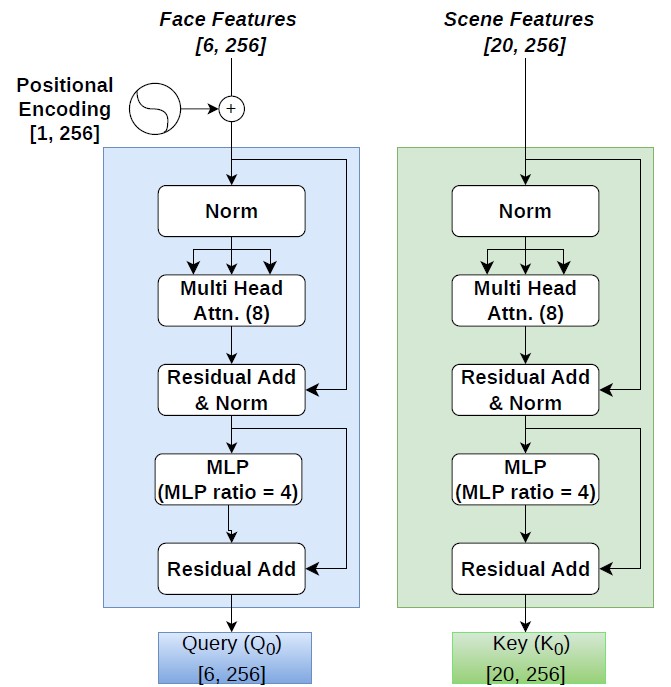}
     \caption{Architecture of encoder block of face features and object scene features embedding representation.}    \label{fig:Encoder_Architecture}
\end{figure}

 The facial feature $F$, given in Equation~\ref{Eq:Facial_Features}, and scene features $S^{obj}$ given in Equation~\ref{Eq:Scene_Features} contain the raw features extracted independently from face and scene objects. These features do not associate with each other contextually. Since gaze estimation is inherently a context-dependent problem, the gaze direction is not determined solely by a single facial region but rather emerges from the interaction among multiple cues, such as eye orientation, head pose, and overall facial geometry. 
Similarly, scene objects are not independent; their spatial arrangement and relative importance influence their likelihood of being the gaze target. To capture this contextual relationship, the face and scene features are passed through two separate standard Transformer encoder blocks \citep{vaswani2017attention}. The flow chart of these encoder blocks is shown in Figure~\ref{fig:Encoder_Architecture}. The outputs of these encoder blocks for face and scene features are treated as queries and keys.

\textit{Limitations of direct feature usage:}  
If raw features are used directly without encoding, each token is treated independently.
% \begin{equation}
% Q = F, \quad K = S
% \end{equation}
In this case, there is no interaction among facial regions and among scene objects. Consequently, the model fails to capture: (a) Relationships between eye regions and head pose (b) Interactions between multiple objects in the scene.

% \begin{itemize}
% \item Relationships between eye regions and head pose
% \item Interactions between multiple objects in the scene
% %\item Contextual dependencies require for accurate gaze estimation
% \end{itemize}

\textit{Role of transformer encoder:}  
The transformer encoder addresses this limitation through self-attention. Each token is updated by attending to all other tokens. So the updated features of face and scene are given below.
\begin{equation}
f_i' = f_i + \sum_{j=1}^{T_f} \alpha_{ij} f_j
\end{equation}
\begin{equation}
s_i^{{obj}'} = s_i^{obj} + \sum_{j=1}^{{N+1}} \beta_{ij} s_j^{obj}
\end{equation}
where $\alpha_{ij}$ and $\beta_{ij}$ are attention weights. This mechanism enables: (a) Each facial token (including 4 face and 2 eye tokens) to incorporate information from all other facial regions (b) Each object feature to capture contextual relationships with other objects.

% \begin{itemize}
% \item Each facial token (including 4 face and 2 eye tokens) to incorporate information from all other facial regions
% \item Each object feature to capture contextual relationships with other objects
% \end{itemize}

%\textbf{Interpretation:}

\textit{Face feature encoding:}  
For facial features, self-attention allows the model to learn dependencies such as (a) alignment between left and right eye features (b) influence of head orientation on gaze direction, and (c) global facial structure. This results in a context aware representation of gaze cues.

% \begin{itemize}
% \item Alignment between left and right eye features
% \item Influence of head orientation on gaze direction
% \item Global facial structure
% \end{itemize}
% This results in a context-aware representation of gaze cues.

\textit{Scene feature encoding:} For scene features, self-attention enables (a) modeling spatial relationships between objects, (b) understanding object grouping and relative importance, and (c) capture contextual interaction in complex driving environments.
% \begin{itemize}
% \item Modeling spatial relationships between objects
% \item Understanding object grouping and relative importance
% \item Capturing contextual interactions in complex driving environments
% \end{itemize}

The encoded features of face and scene are represented as:
\begin{equation}
Q_0 = \text{Encoder}(F), \quad
K_0 = \text{Encoder}(S^{obj}) \quad
%V_0 = K_0
\end{equation}

% where $Q_0 = \begin{bmatrix} q_1 \\ q_2 \\ \vdots \\ q_{T_f} \end{bmatrix}$ and $Q = \begin{bmatrix} k_1 \\ k_2 \\ \vdots \\ k_N \end{bmatrix}$

\begin{equation*}
Q_0 = \begin{bmatrix}
q_{0,1} \\
q_{0,2} \\
\vdots \\
q_{0,T_f}
\end{bmatrix}, \quad
K_0 = \begin{bmatrix}
k_{0,1} \\
k_{0,2} \\
\vdots \\
k_{0,N}
\end{bmatrix}
\end{equation*}

where $q_{0,1}, q_{0,2}, \ldots, q_{0,T_f}$ represent the elements of the query vector $Q_0$, and $k_{0,1}, k_{0,2}, \ldots, k_{0,{N+1}}$ represent the elements of the key vector $K_0$.

% =========================================================
%\subsubsection{Query-Key-Value Projection}

After obtaining the encoded representations ($Q_0$) and ($K_0$) from the Transformer encoders, linear projection layers are applied to generate the query and key matrices, defined as: 

%Linear projections are applied to transform the encoded features:

\begin{equation}
Q = Q_0 W_q = \begin{bmatrix}
q_1 \\
q_2 \\
\vdots \\
q_{T_f}
\end{bmatrix}, \quad K = K_0 W_k = \begin{bmatrix}
k_1 \\
k_2 \\
\vdots \\
k_{N+1}
\end{bmatrix} \quad %V = V_0 W_v
\label{Query-Keys}
\end{equation}

% \begin{equation*}
% Q = \begin{bmatrix}
% q_1 \\
% q_2 \\
% \vdots \\
% q_{T_f}
% \end{bmatrix}, \quad
% K = \begin{bmatrix}
% k_1 \\
% k_2 \\
% \vdots \\
% k_{N+1}
% \end{bmatrix}
% \end{equation*}

\par Although the encoder outputs capture rich contextual information within the driver facial features and scene features, they are not inherently optimized for cross modal matching between facial gaze cues and scene objects. The projection step addresses this limitation by transforming the encoded features into a task-specific embedding space, where similarity can be effectively computed through dot product attention. In particular, the projection matrices ($W_q$) and ($W_k$) learn how to reorient the feature representations such that gaze relevant patterns in facial features align with corresponding object features in the scene. Consequently, the attention score can be interpreted as a learned similarity function, rather than a simple dot product between raw features. This transformation is help for improving the discriminative ability of the model, especially in scenarios involving visually similar or spatially proximate objects, thereby enhancing the accuracy of gaze object prediction.

% =========================================================
\subsection{Attention between encoded facial and scene features}

The interaction between queries (Q) derived from facial features and keys (K) derived from scene objects using Equation~\ref{Query-Keys} is used to compute similarity scores. These similarity scores are calculated using the scaled dot-product attention mechanism, as given in Equation~\ref{scaled_dot}.

% \begin{equation}
% S = \frac{QK^T}{\sqrt{d}}
% \end{equation}

\begin{equation}
S = \frac{QK^T}{\sqrt{d}}
\label{scaled_dot}
\end{equation}

where, $S = \{s_{i,j}\} \in \mathbb{R}^{T_f \times (N+1)}$ is the attention score matrix, and each element $s_{i,j}$ is defined as $s_{i,j} = \frac{q_i \cdot k_j}{\sqrt{d}}$, where $q_i$ and $k_j$ denote the $i$-th query vector and $j$-th key vector, respectively.

% where, $S = \{s_{i,j}\} \in \mathbb{R}^{T_f \times (N+1)}$ is the attention score matrix, and each element $s_{i,j}$ is defined as
% \begin{equation}
% s_{i,j} = \frac{q_i \cdot k_j}{\sqrt{d}},
% \end{equation}

% where, $q_i$ and $k_j$ denote the $i$-th query vector and $j$-th key vector, respectively.

% \begin{equation*}
% s_{i,j} = \frac{q_i \cdot k_j}{\sqrt{d}}
% \end{equation*}
% where, $q_i$ and $k_j$ denote the $i$-th query vector and $j$-th key vector, respectively.

\textbf{\textit{Masking on zero-padded objects:}} In the scene image, if the number of detected traffic objects is lesser than 19, virtual objects are introduced to maintain a fixed input size. These additional objects are treated as zero-padded objects, where all elements of their corresponding feature vectors are set to zero.

During the attention score computation, a masking mechanism is applied to these zero padded (virtual) objects. Specifically, the attention scores corresponding to such objects are assigned a value of $-\infty$. 
Virtual objects (padding) are masked using the following equation:
\begin{equation}
s_{ij} =
\begin{cases}
s_{ij}, & \text{valid} \\
-\infty, & \text{invalid}
\end{cases}
\end{equation}

Consequently, after applying the softmax function, the attention weights for these virtual objects become zero, ensuring that they do not contribute to the final attention output.

The attention weights are computed using temperature scaling:
\begin{equation}
A = \text{softmax}\left(\frac{S}{\tau}\right)
\end{equation}
% where $\tau = 0.5$.

To compute attention weights, a temperature parameter $\tau = 0.5$ is applied to the similarity scores before the softmax operation. This scaling controls the sharpness of the attention distribution by preventing excessively large values from dominating the softmax output. Without temperature scaling, the attention weights can become overly peaked, leading to unstable gradients and poor generalization. By introducing a temperature parameter, the model can control how spread out the attention over the objects. This helps the model focus more on the most important objects while still considering other possible objects, which improves learning stability of the model.

% =========================================================
\textbf{\textit{Face-eye attention fusion:}}The attention weights are separated into face and eye components because driver gaze is depends on head orientation and iris position within the eye or eye movement. Eye features provide precise information about gaze direction but can be sensitive to noise and occlusions, whereas face features capture head orientation and offer more stable but coarse cues of driver gaze direction. By separating face and eye attention weights the model learn fine grained eye information and maintaining robustness through face-based context.

\begin{equation}
A_{\text{face}} = \frac{1}{4} \sum_{i=1}^{4} A_i
\end{equation}

\begin{equation}
A_{\text{eye}} = \frac{1}{2} \sum_{i=5}^{6} A_i
\end{equation}

The final attention weights is a combination of weighted average of attention weights of eye and the face components over objects, where learnable parameter $\lambda$ decide the contribution of each components.
\begin{equation}
A_{\text{final}} = \lambda A_{\text{eye}} + (1 - \lambda) A_{\text{face}}
\end{equation}

The higher the $\lambda$ value model giving more weights to eye component and vice versa.

The final attention weight matrix $A_{\text{final}}$ represents the degree of attention assigned by the model to each object in the scene. It is defined as:
\begin{equation}
A_{\text{final}} = [a_1, a_2, \dots, a_{N+1}] \in \mathbb{R}^{{N+1}}
\end{equation}
where $a_i \in [0,1]$ denotes the attention weight corresponding to the $i$-th object, and $\sum_{i=1}^{{N+1}} a_i = 1$. After computing the attention weights between the queries and keys coming from the facial features and scene features respectively , then we have to compute the object index id the details of which is given next section.

\subsection{Gaze estimation head}

\subsubsection{Gaze object prediction}
The final attention weight matrix is used to compute the gaze object. The object for which the computed attention weight is highest is considered the gaze object (i.e., the object at which the driver is looking).
%The final attention weight vector $A_{\text{final}}$ directly represents the prediction scores over all objects:
% \begin{equation}
%  A_{\text{final}} = [a_1, a_2, \dots, a_N] \in \mathbb{R}^{N}
% \end{equation}

% where $a_j \in [0,1]$ denotes the attention weight corresponding to the $j$-th object, and $\sum_{j=1}^{N} a_j = 1$.
% This can be mathmatically formulated as
% Then the estimated gaze object out of the total number of object present in the scene is:
This can be mathematically formulated as:
\begin{equation}
\hat{y} = \arg\max_{j} a_j
\end{equation}

\noindent where $\hat{y}$ denotes the predicted gaze object, represented as the index of the object with the highest attention weight, i.e., $\hat{y} \in \{1, 2, \dots, {N+1}\}$.

% \noindent where $\hat{y}$ represents the estimated gaze target.

% =========================================================
\subsubsection{Loss functions}
To effectively train the proposed gaze estimation model, multiple loss components are employed to guide different aspects of learning, including classification accuracy, attention alignment, consistency, and discrimination among similar objects. Each component of the loss function is discussed in detail below.

%\subsubsection{Classification Loss}
\textbf{\textit{Classification loss:}}
The primary objective is to correctly estimate the gaze target among $N$ traffic objects and background. This is formulated as a multi class classification problem using cross entropy loss:

% \begin{equation}
% \mathcal{L}_{cls} = - \log \left( \frac{\exp(z_y)}{\sum_j \exp(z_j)} \right)
% \end{equation}

\begin{equation}
\mathcal{L}_{cls} = - \log \left( \frac{\exp(a_y)}{\sum_{j=1}^{{N+1}} \exp(a_j)} \right)
\end{equation}

where $a_j$ denotes the estimated attention weight for the $j$-th object, and $a_y$ is the ground truth object attention weight. Classification loss ensures that the model assigns the highest probability to the correct object. It serves as the primary supervision signal for gaze object prediction.

% -------------------------
\textbf{\textit{Consistency loss:}}
This component of loss function ensure coherence between eye and face based attention, a consistency constraint is introduced:
\begin{equation}
\mathcal{L}_{cons} = \alpha \|A_{\text{eye}} - A_{\text{face}}\|_2^2
\end{equation}
%where $\alpha = 0.03$.

%\textbf{Explanation:}  
where $\alpha = 0.03$. This enforces agreement between eye-based and face-based attention, reducing inconsistent predictions.
This allows the model to automatically balance the contributions of eye and face components, avoiding manual tuning and improving optimization.

% -------------------------
\textbf{\textit{Confusion-aware attention loss:}}
In complex driving scenes, multiple traffic objects are often located close to one another, making gaze-object prediction inherently ambiguous. Conventional one hot supervision treats all incorrect objects equally, thereby over penalizing predictions on nearby objects that are more likely to be confused with the true gaze target. Therefore, a confusion-aware supervision strategy is introduced to explicitly model this spatial ambiguity.
\par The proposed approach constructs a soft target distribution based on the spatial distances between the ground truth object and all other detected objects, assigning higher probabilities to nearby objects. This distance-aware distribution is combined with the one-hot ground-truth label, and the predicted eye-attention distribution is optimized using KL divergence \citep{kullback1951information} to learn spatially aware attention while preserving strong supervision for the correct gaze object.

\par The distance between objects is computed using the Euclidean distance between their center coordinates. Specifically, for the $i$-th object, the distance from the ground truth object is defined as:

%To account for spatial ambiguity and nearby object confusion, a soft target distribution is constructed based on the distance between objects. 
\begin{equation}
ED^{obj}_i = \sqrt{(x_i - x_{gt})^2 + (y_i - y_{gt})^2}
\end{equation}

where $(x_i, y_i)$ and $(x_{gt}, y_{gt})$ denote the center coordinates of the $i$-th object and the ground truth object, respectively.

% \begin{equation}
% p_i = \frac{\exp(-\gamma ED^{obj}_i)}{\sum_j \exp(-\gamma ED^{obj}_i)}
% \end{equation}
% where $\gamma$ is hyperparameter that controls how distance is converted into probability.

This is converted into a probability distribution using a softmax function over the negative distances:
\begin{equation}
p_i=\frac{\exp\!\left(-\gamma ED_i^{\mathrm{obj}}\right)}
{\sum_{j=1}^{N+1}\exp\!\left(-\gamma ED_j^{\mathrm{obj}}\right)},
\end{equation}
where $p_i$ denotes the probability associated with the $i$-th object, $\gamma$ is a scaling hyperparameter that controls the sharpness of the resulting probability distribution. Larger values of $\gamma$ assign higher probabilities to objects closer to the predicted gaze point.

The final target distribution is:
% \begin{equation}
% g = 0.9 \cdot \text{onehot}(y) + 0.1 \cdot p
% \end{equation}
\begin{equation}
g = 0.9 \cdot \mathrm{OneHot}(y) + 0.1 \cdot p
\end{equation}
where $y$ is the ground truth object index which belongs $y \in \{1, 2, \dots, {N+1}\}$ 

Then the confusion aware attention loss is computed using following Equation:
\begin{equation}
\mathcal{L}_{attn} = D_{KL}(  g \parallel A_{\text{eye}})=\sum_{j=1}^{{N+1}} g_j \log \frac{g_j}{A_{\text{eye},j}}
\end{equation}

% \begin{equation}
% \mathcal{L}_{attn} = D_{KL}(  g \parallel A_{\text{eye}})= \sum_{j} g_j \log \frac{g_j}{A_{\text{eye},j}} 
% \end{equation}

% where g is a soft target distribution. This loss encourages the model to assign the highest probability to the correct object while also allocating smaller probabilities to nearby or contextually relevant objects, thereby improving robustness in ambiguous or uncertain scenarios.
The Kullback–Leibler (KL) divergence is used to measure the difference between the predicted attention distribution and the target distribution, where $g_j$ denotes the ground-truth probability assigned to the $j$-th object, and $A_{\text{eye},j}$ represents the predicted attention weight for the $j$-th object. Here, ${N+1}$ is the total number of objects in the scene. Minimizing this divergence encourages the predicted attention distribution to align closely with the target distribution.

% -------------------------
\textbf{\textit{Hard negative margin loss:}}
Although the confusion aware attention loss encourages the model to assign higher attention to the ground truth object and nearby objects, it does not explicitly enforce sufficient separation between the ground truth object and the most confusing incorrect objects. In driver gaze object prediction, these confusing objects, referred to as hard negatives, are those that receive high attention scores despite not being the ground truth object. %Hard negatives typically correspond to traffic objects that are spatially close to the gaze target, making them difficult for the model to distinguish.

\par To reduce this ambiguity, a margin based loss is introduced that explicitly increases the gap between the attention score of the ground truth object and those of the hardest negative objects. During training, the top-k non ground truth objects with the highest attention scores are selected as hard negatives. The model is then encouraged to maintain a predefined margin between the attention assigned to the ground truth object and the average attention of these hard negatives, thereby improving discriminative learning.

%This loss reduce confusion between visually similar or nearby objects, a margin-based loss is applied:

\begin{equation}
\mathcal{L}_{margin} = \max(0, m - (a_{gt} - a_{neg}))
\end{equation}

% where:
% \begin{itemize}
% \item $m = 0.5$
% \item $a_{gt}$ is the score of the ground truth object
% \item $a_{neg}$ is the mean of top-2 competing object scores
% \end{itemize}

\noindent  where $a_{gt}$ denotes the attention weight corresponding to the ground truth object, and $a_{neg}$ denotes the average attention weight of the top-$k$ (k=2 in our case) hardest negative objects. $m = 0.5$, is the minimum required difference between them.

%\textbf{Explanation:}  
%This loss pushes the model to create a margin between the correct object and hard negative objects that are visually or spatially similar.

The margin loss is active only when the difference between the attention assigned to the ground-truth object and the hard negatives is smaller than the predefined margin $m$. In this case, the loss penalizes the model and encourages it to increase the attention of the ground-truth object while suppressing the attention of the competing hard negatives. Once the required margin is achieved, the loss becomes zero, preventing unnecessary optimization. Consequently, the model learns more discriminative attention representations, reducing confusion between the true gaze object and visually or spatially similar objects while improving gaze object prediction  performance.

% -------------------------
\textbf{\textit{Final loss:}}
The final loss consists of classification loss, consistency loss, confusion aware attention loss, hard negative margin loss to train the proposed gaze estimation model effectively.

%To effectively train the proposed gaze estimation model, multiple loss components are employed to guide different aspects of learning, including classification accuracy, attention alignment, consistency, and discrimination among similar objects.
\begin{equation}
\mathcal{L} =
\mathcal{L}_{cls}
+ 0.5 \mathcal{L}_{margin}
+ e^{-\sigma_1} \mathcal{L}_{cons} + \sigma_1
+ e^{-\sigma_2} \mathcal{L}_{attn} + \sigma_2
\end{equation}

The consistency loss and confusion aware attention loss capture different aspects of the proposed gaze estimation framework and may exhibit different optimization characteristics during training. Assigning fixed weights to these loss terms requires manual tuning and may not provide an optimal balance throughout the training process. Therefore, an uncertainty based weighting strategy is adopted, in which the contribution of each loss is automatically determined through learnable uncertainty parameters, $\sigma_1$ and $\sigma_2$.

Each loss term is weighted by an exponential factor, $e^{-\sigma_1}$/$e^{-\sigma_2}$, such that losses associated with higher uncertainty receive lower weights, whereas more reliable losses contribute more strongly to the overall optimization. However, using only the weighting terms $e^{-\sigma_1}\mathcal{L}_{cons}$ / $e^{-\sigma_2}\mathcal{L}_{attn}$ would allow the optimization to trivially increase $\sigma_1$/$\sigma_2$, thereby driving the corresponding loss weights toward zero and effectively removing these loss terms from the training objective. To avoid this degenerate solution, an additional regularization term, $+\sigma_1$/$+\sigma_2$, is included for each uncertainty parameter. This term penalizes excessively large uncertainty values, forcing the optimization to learn an appropriate trade off between reducing the weighted loss and keeping the uncertainty bounded.

\subsubsection{Evaluation metric}
\label{evaluation_metrics}
The model is evaluated in terms of accuracy, which is defined as the proportion of samples for which the estimated gaze object index matches the ground-truth index. Formally, for a dataset containing $M$ samples in the testing, accuracy is computed as
% \[
% \mathrm{Accuracy} = \frac{1}{M} \sum_{i=1}^{M} \mathbb{I}\left( \hat{y}_i = y_i \right),
% \labe{Accuray}
% \]
\[
\mathrm{Accuracy} = \frac{1}{M} \sum_{i=1}^{M} \mathbb{I}\left( \hat{y}_i = y_i \right)
\label{eq:accuracy}
\]
where $y_i$ and $\hat{y}_i$ denote the ground truth and predicted gaze object indices for the $i$-th sample, respectively, and $\mathbb{I}(\cdot)$ is the indicator function.

%=========================================================
\subsection{Training details}
The proposed model is trained using the PyTorch framework on a GPU server with four NVIDIA GeForce RTX 3080 GPUs (10 GB VRAM each) and CUDA 11.4. The details of the dataset used for the training, validation and testing is discussed in Table \ref{tab:Train_Val_Test}. The total number of data samples consists of 165,969, 9,208, and 14,673 images in the training, validation, and testing sets, respectively. The model is optimized using the AdamW optimizer with an initial learning rate of $1 \times 10^{-5}$ and weight decay to improve generalization. A cosine annealing learning rate scheduler is used to gradually reduce the learning rate during training, ensuring stable convergence. The model is trained using mini-batches with a batch size of 64, and gradient clipping is applied with a maximum norm of 1.0 to prevent exploding gradients.

\section{Results}
In this section, we first evaluate the performance of the proposed Transformer based Gaze Object (TransGaze-Object) prediction model. This is followed by a comparison of our proposed model's results with the existing state-of-the-art point-of-gaze estimation model, SGAP-Gaze \citep{sharma2026sgap}, by associating the point of gaze with the object bounding box via post-processing. Finally, we discuss the incorrect gaze-object prediction analysis of the proposed model.

\subsection{Overall accuracy}
\begin{figure}[!htbp]
    \centering
    \includegraphics[width=0.95\textwidth]{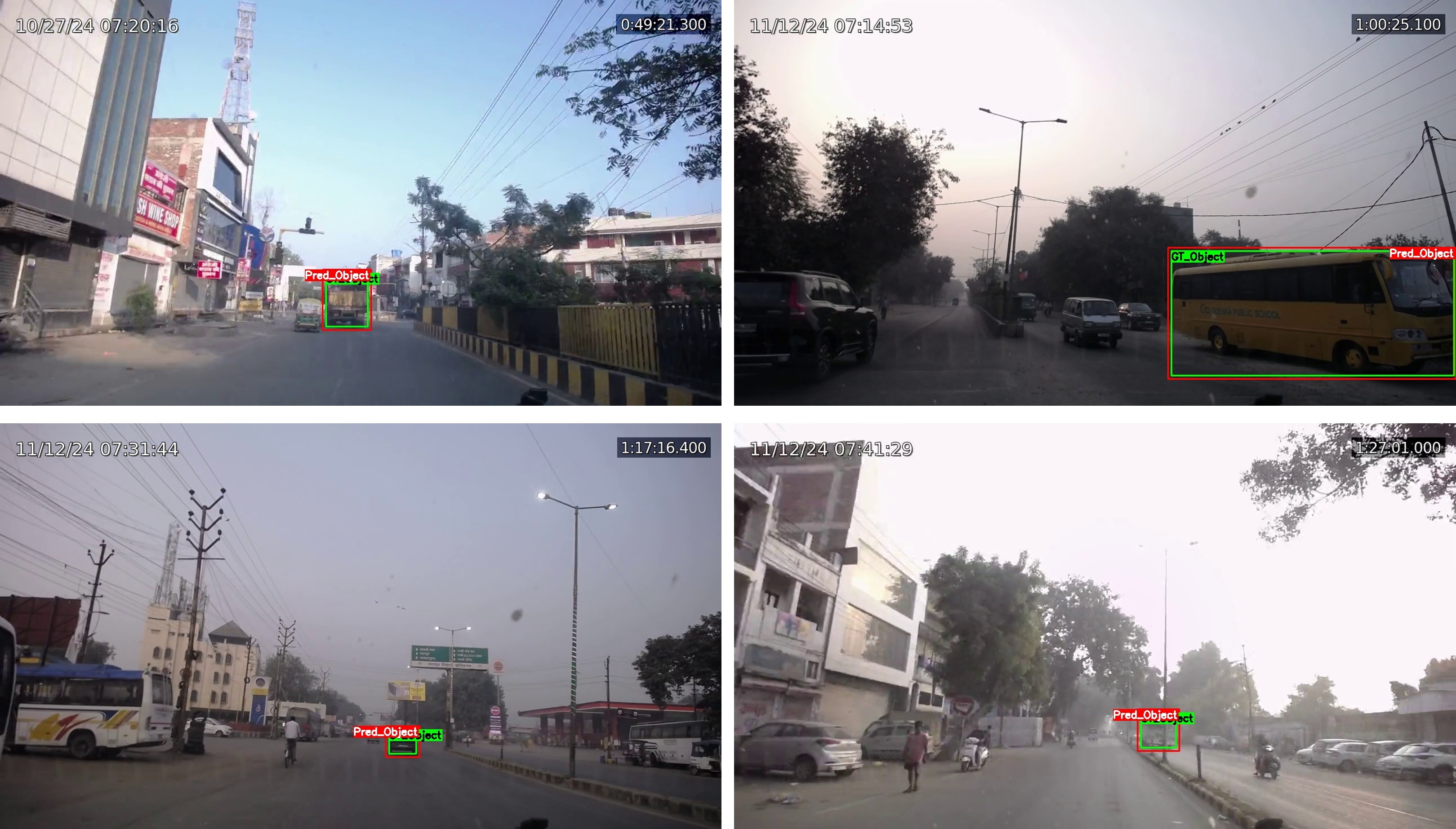}
     \caption{Test samples of correct predicted gaze object, indicating green bounding box is ground truth and red is predicted gaze object.}    \label{fig:Correct_Estimated_Gaze_Object}
\end{figure}

The model performance was evaluated on three different drivers, 14,673 test samples. The test data are completely different from the data used to train and validate the model. The model's accuracy was evaluated by predicting gaze objects on 14,673 test samples, of which 8,723 were correctly predicted. This results in an overall gaze-object prediction accuracy of 59.45\%, computed using Equation ~\ref{eq:accuracy}. 
The representative test samples of correctly predicted gaze objects are shown in Figure~\ref{fig:Correct_Estimated_Gaze_Object}. In each sample, the red and green bounding boxes denote the predicted and ground-truth gaze objects, respectively. Since the prediction is correct, both bounding boxes represent the same traffic object.

\subsection{Performance comparison of TransGaze-Object and SGAP-Gaze}

The performance of our proposed TransGaze-Object prediction model has been compared with the PoG model by associating the estimated gaze point with object bounding boxes. Please note that, to the best of the author's knowledge, no gaze-object prediction model exists in the literature that can directly represent gaze in terms of the object. Therefore, we have used our previously proposed state-of-the-art PoG model (SGAP-Gaze) \citep{sharma2026sgap} for comparison by associating the gaze point with the object's bounding box detected in a scene image via post-processing. If the gaze point lay inside a bounding box, the corresponding object was assigned as the gaze object.

\par For performance evaluation, both the SGAP-Gaze and TransGaze-Object models were evaluated using the same test dataset. First, the point of gaze for the test samples was estimated using the SGAP-Gaze model, and the resulting gaze points were associated with the corresponding detected objects' bounding boxes in the scene image. If the estimated point of gaze lies within an object's bounding box and corresponds to an object that is a ground-truth object, then it is assigned a value of 1, indicating a correctly predicted gaze object; otherwise, it is assigned a value of 0. The association through the point-of-gaze model achieved an accuracy of 50.98\%. In contrast, the proposed TransGaze-Object model achieves an accuracy of 59.45\%, representing an overall 8.47\% improvement compared to SGAP-Gaze, which confirms that incorporating bounding-box information improves gaze-object prediction accuracy.   
However, our proposed model accuracy remains lower, with only about 60\% of gaze objects correctly predicted. Given that even state-of-the-art PoG-based gaze object prediction has achieved an accuracy of only around  51\%, this finding highlights that gaze object prediction is a significantly challenging problem, which is often not reflected by the small values of angular error ($\sim 6^\circ$) reported in the gaze direction estimation literature \citep{kasahara2022look, cheng2024you}. Next, to investigate the possible reasons of incorrect predictions, we analyze the factors contributing to gaze object prediction errors.

%In the next, we analyses the possible reasons of failure of correct gaze object prediction of our TransGaze-Object prediction model. 

\subsection{Error analysis of gaze object predictions}
%To understand the failure cases, false predicted gaze object are categories in three categories. 
The predicted incorrect gaze object is categorized into three groups to understand the possible reasons of the failure cases. In the first category, both the predicted and ground-truth gaze objects are traffic objects, but they correspond to different object IDs. %The traffic object can be any of 10 classes, such as pedestrian, rider, bicycle, motorcycle, auto-rickshaw, car, bus, truck, traffic sign, or traffic light. 
In the second category, the predicted gaze object is the background, whereas the ground-truth is one of the traffic objects in the traffic scene. And finally, in the third category, the predicted gaze object is a traffic object, while the background is the ground truth. The results of all three failure cases are shown in Table \ref{tab:Comparison of results of TransGaze-Object and PoG Mapping}.

In the first category, where both the ground-truth and incorrectly predicted gaze objects correspond to traffic objects, the performances of the TransGaze-Object model and SGAP-Gaze are nearly identical. As shown in Table~\ref{tab:Comparison of results of TransGaze-Object and PoG Mapping}, the corresponding incorrect predictions are 16.23\% and 16.55\%, respectively. It should be noted that these percentages are calculated with respect to the total number of test samples (14,673).

In the second category, where the ground truth gaze object is a traffic object but the predicted gaze object is the background, the proposed TransGaze-Object model demonstrates a significant improvement. TransGaze-Object produces only 11.68\% incorrect predictions, whereas SGAP-Gaze (PoG associated to objects) incorrectly predicted gaze object 23.21\% of the total test samples as background. These results indicate that incorporating object-level spatial information enables the proposed TransGaze-Object model to distinguish traffic objects from the background more effectively, reducing this type of error by nearly half compared to SGAP-Gaze.

In the third category, where the ground truth gaze object is the background but the predicted gaze object is a traffic object, the performance of TransGaze-Object is slightly inferior to that of SGAP-Gaze. The corresponding error rates are 12.62\% and 9.24\%, respectively.
This suggests that although the proposed model is more effective at reducing the incorrect prediction of traffic objects as background gaze objects, it exhibits a slight increase in the incorrect prediction of background as a traffic object. However, overall, the performance improvement in TransGaze-Object model is primarily contributed to reducing the incorrect prediction from background to traffic object (category 2), justifying the usefulness of providing the bounding box information apriori.

\begin{table*}[ht]
\centering
\label{tab:False_Gaze_Object_Estimation}
\begin{doublespace}
\caption{Comparison of false predicted gaze object of TransGaze-Object and PoG-to-object association using SGAP-Gaze}
\label{tab:Comparison of results of TransGaze-Object and PoG Mapping}
\begin{tabular}{lcccc}
\hline
\multirow{2}{*}{Categories}   & \multicolumn{2}{l}{TransGaze-Object} & \multicolumn{2}{l}{PoG-Object Association} \\ \cline{2-5} 
                              & Count          & Percentage(\%)           & Count            & Percentage(\%)            \\ \hline
Traffic Object-Traffic Object & 2382           & 16.23               & 2429             & 16.55                 \\
Traffic Object-Background     & 1715           & 11.68               & 3407             & 23.21                 \\
Background-Traffic Object     & 1853           & 12.62               & 1357             & 9.24                  \\ \hline
\end{tabular}
\end{doublespace}
\end{table*}

\par
\begin{figure}[!htbp]
    \centering
    \includegraphics[width=0.95\textwidth]{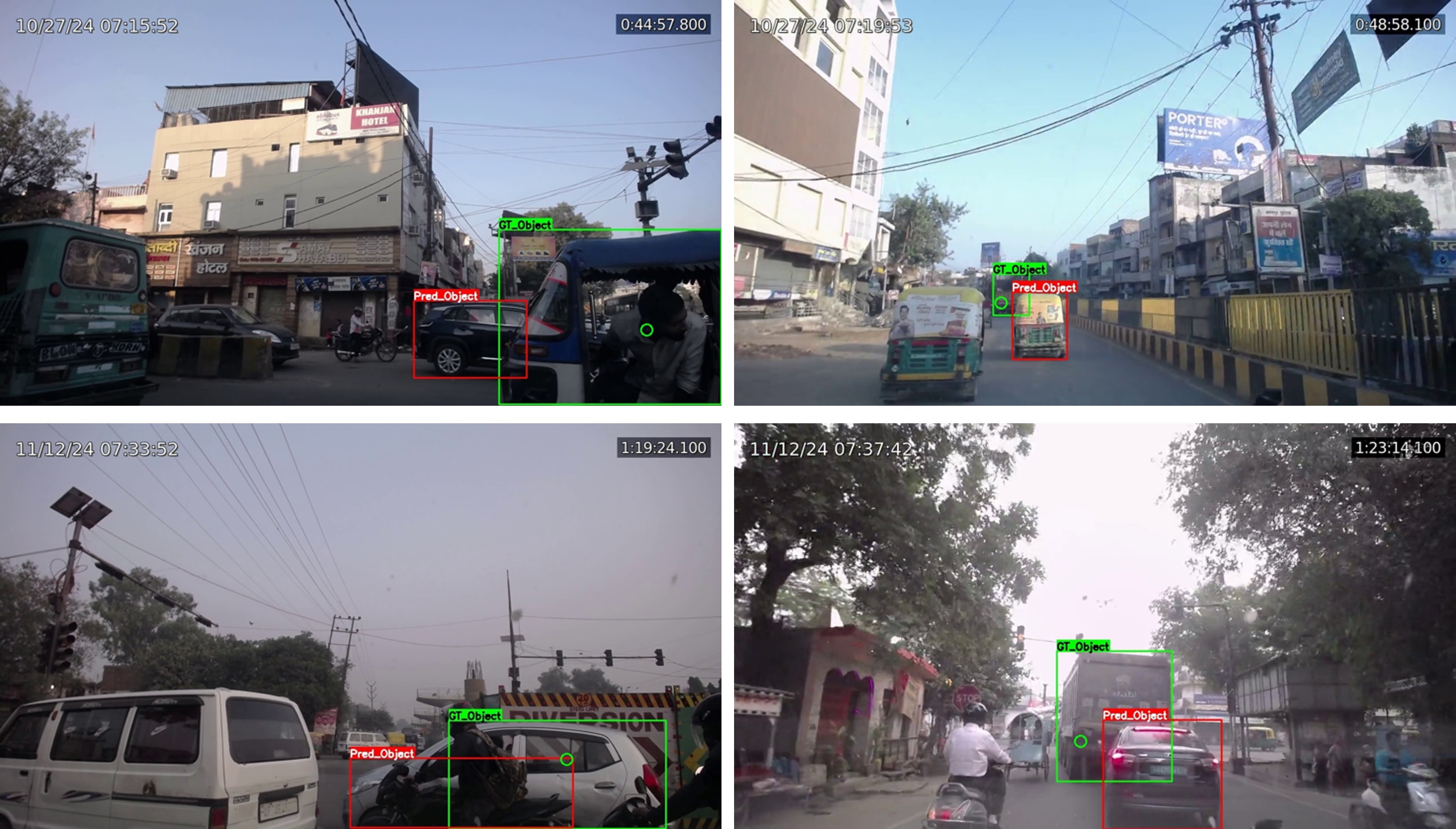}
     \caption{Illustration of test samples showing overlapping traffic objects with incorrectly predicted gaze object }    \label{fig:Misclassified_Overlapping_Gaze_Object}
\end{figure}

 A detailed analysis of incorrect gaze object predictions, in which both the ground truth and the predicted objects are traffic objects, is presented in Table~\ref{Overlapping-NonOverlapping}. These incorrect gaze object predictions are further categorized as overlapping or non-overlapping traffic objects based on the ground truth and predicted object bounding boxes. A incorrect gaze object prediction is considered overlapping, if the ground truth and predicted bounding boxes share  common area (i.e., have a non-zero intersection), otherwise, it is categorized as a non-overlapping. As shown in the Table~\ref{Overlapping-NonOverlapping}, the proposed TransGaze-Object model exhibits 353 (2.40\%) overlapping traffic object to traffic object incorrect gaze object prediction, which is slightly higher than 331 (2.25\%) observed for SGAP-Gaze. These errors mainly occur when multiple traffic objects are located close to each other and have overlapping bounding boxes, making them visually difficult to distinguish. Figure~\ref{fig:Misclassified_Overlapping_Gaze_Object} shows a sample of the incorrect prediction of gaze object due to the overlapping of ground truth and predicted traffic objects. 
\par
\begin{figure}[!htbp]
    \centering
    \includegraphics[width=0.95\textwidth]{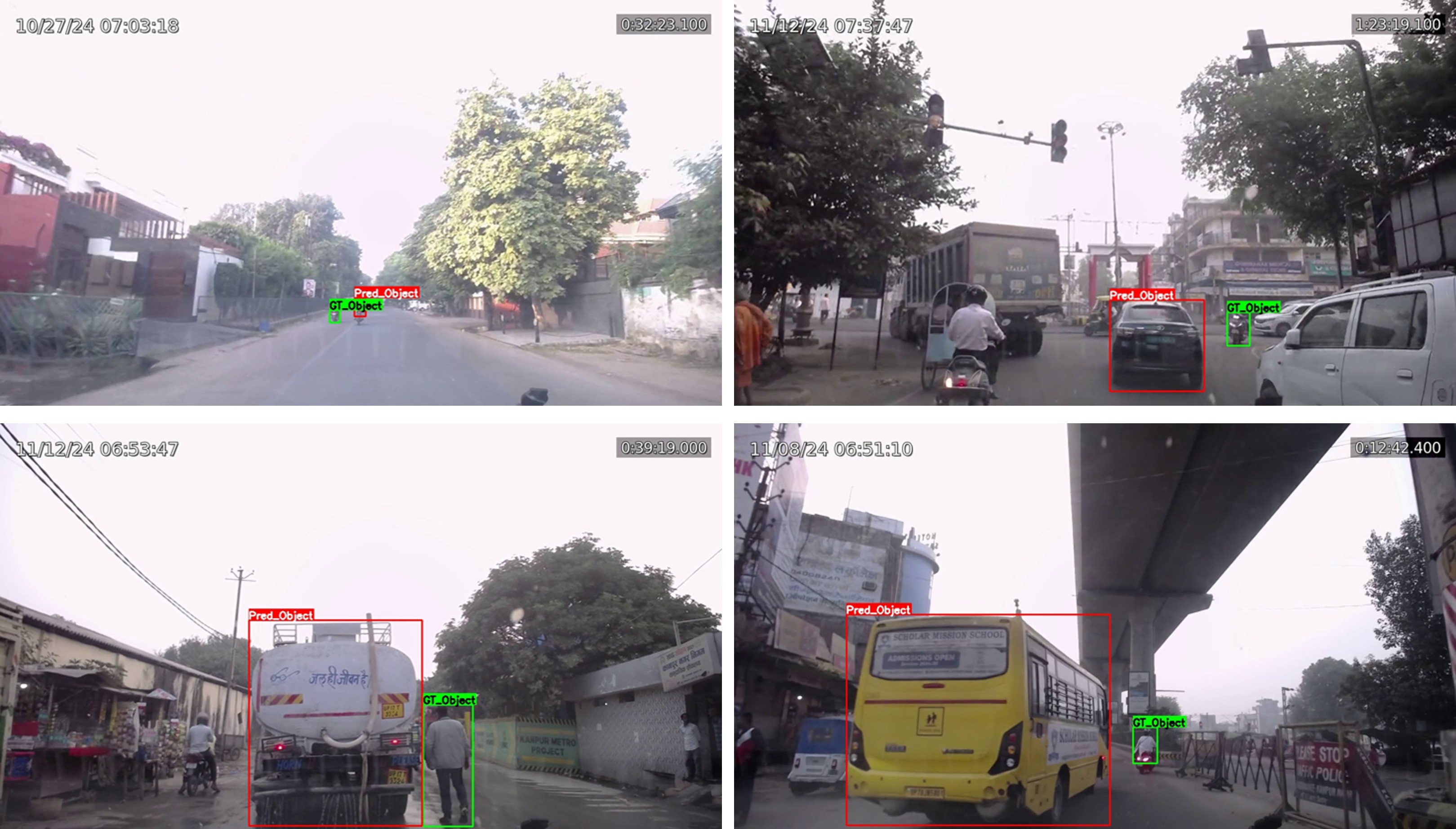}
     \caption{Test samples of false predicted gaze objects (Non-overlapping)}    \label{fig:Misclassified_Non_Overlapping_Gaze_Object}
\end{figure}

 For non-overlapping traffic object to traffic object incorrect prediction, TransGaze-Object records 2029 (13.82\%) incorrect samples, whereas SGAP-Gaze (PoG-object association) records 2098 (14.29\%) incorrect samples. The slightly lower error rate of TransGaze-Object indicates that the proposed model is marginally more effective in discriminating between spatially separated traffic objects. Figure~\ref{fig:Misclassified_Non_Overlapping_Gaze_Object} shows a sample of the false prediction of gaze object due to the non-overlapping of ground truth and predicted traffic objects. Overall, the false prediction rates for traffic object-to-traffic object comparisons of the two methods (TransGaze-Object and SGAP-Gaze) remain comparable, suggesting that the primary advantage of the proposed TransGaze-Object model lies in reducing confusion between traffic objects and the background rather than between different traffic objects.

% \usepackage{multirow}
\begin{table*}[ht]
\centering
\label{tab:False_Gaze_Object_Estimation}
\begin{doublespace}
\caption{Analysis of incorrect gaze object prediction between traffic object - traffic object}
\label{Overlapping-NonOverlapping}
\begin{tabular}{lcccc}
\hline
\multirow{2}{*}{Categories}   & \multicolumn{2}{l}{TransGaze-Object} & \multicolumn{2}{l}{PoG-Object Association} \\ \cline{2-5} 
                              & Count          & Percentage(\%)          & Count            & Percentage(\%)            \\ \hline
\begin{tabular}[c]{@{}l@{}}Traffic Object-Traffic Object\\ (Overlapping)\end{tabular}     & 353            & 2.40                & 331              & 2.25                  \\
\begin{tabular}[c]{@{}l@{}}Traffic Object-Traffic Object\\ (Non-Overlapping)\end{tabular} & 2029           & 13.82               & 2098             & 14.29                 \\ \hline
\end{tabular}
\end{doublespace}
\end{table*}

% \subsubsection{Error Analysis of Traffic Gaze Object Predictions vs. Ground Truth Background}
%\subsubsection{Error analysis of traffic gaze object predictions with background ground truth}
\subsection{Error analysis of background and traffic object}
\par To investigate the effect of object scale on gaze estimation performance, we analyzed the relationship between the predicted bounding-box area and gaze estimation error. The estimated error is computed as the shortest distance between the ground truth point-of-gaze coordinates and the nearest edge of the predicted gaze object bounding box.
The bounding box coordinates, originally in normalized form, are first converted to pixel coordinates using the image resolution (1280 × 720). The area of each estimated bounding box is then computed in pixel units. To ensure a fair comparison across objects of different sizes, the gaze error is normalized by the square root of the bounding box area, which provides a scale-invariant measure of localization error.

Furthermore, objects are categorized into three groups: small, medium, and large based on their bounding box areas. This categorization is performed using quantile-based partitioning, in which the dataset is divided into three equal subsets based on the distribution of bounding box areas. This ensures a balanced representation of object sizes and allows for meaningful comparison of model performance across different scales. The average normalized error is then computed for each size group to evaluate the influence of object scale on gaze object prediction accuracy.

\begin{figure}[!htbp]
    \centering
    \includegraphics[width=0.95\textwidth]{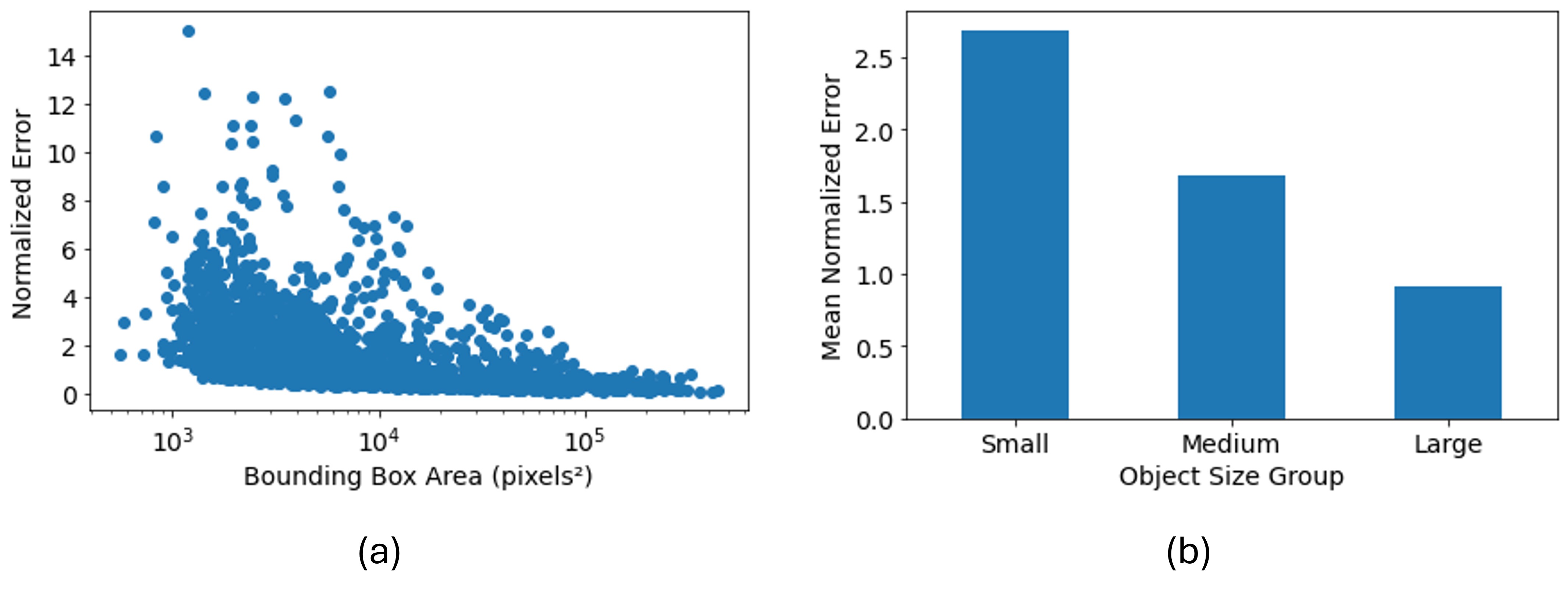}
     \caption{Test samples of failure cases where predicted gaze object is traffic object while ground truth is background}    \label{Mean_Normalized_Error}
\end{figure}

The analysis reveals a clear negative correlation between bounding box area and normalized gaze error (Pearson = -0.286, Spearman = -0.622), also shown in Figure~\ref{Mean_Normalized_Error}a, indicating that larger objects are associated with lower relative error. Additionally, the size-wise evaluation shows that small objects exhibit significantly higher normalized error compared to medium and large objects, as shown in Figure~\ref{Mean_Normalized_Error}b. This suggests that the proposed model performs more reliably on larger objects, while smaller objects remain challenging due to their limited spatial extent and increased ambiguity in gaze-object association. %Figure~\ref{fig:Ground_Truth_Background_Estimated_Gaze_Object} shows test samples of failure cases where the predicted gaze object is a traffic object while the ground truth is the background. Bounding box with green color shows the background which draw over whole image, red is the predicted traffic object, green circle center represent the actual ground truth gaze coordinate obtained from eye tracker data. 

% \begin{figure}[!htbp]
%     \centering
%     \includegraphics[width=0.95\textwidth]{Figures/Ground_Truth_Background_Estimated_Gaze_Object.png}
%      \caption{Failure cases of test samples where predicted gaze object is traffic object while ground truth is background.}    \label{fig:Ground_Truth_Background_Estimated_Gaze_Object}
% \end{figure}
% Overall, the detailed analysis of the TransGaze-Object model shows that predicting the traffic object, while helpful to associate driver attention in a meaningful manner. 
\section{Conclusions}
Driver gaze provides a significant role in assessing driver visual attention and situational awareness. The existing approach of driver gaze estimation represents the driver's gaze either in the interior vehicle region, such as the forward windshield, side wing mirror, or as a gaze direction vector/point of gaze on the scene image. Object-level gaze information provides more semantically meaningful cues for driver visual attention and situational awareness, thereby helping develop gaze-based driver monitoring systems to improve driver safety. However, no existing studies have performed end-to-end gaze object prediction.  
Therefore, in this study, we proposed a transformer-based gaze object (TransGaze-Object) prediction framework that directly represents the driver's gaze as a traffic object or the background. The proposed framework requires face and scene images as inputs to extract facial and scene features. First, the Face-Eye-Iris and traffic objects are detected in the face and scene images, respectively, using two separate custom YOLOv8-based detectors. The proposed framework then extracts facial features, including face and iris-weighted eye features, along with spatial features of the detected traffic objects. A transformer-based cross-attention mechanism is then used to compute similarity scores and attention weights for predicting the driver's gaze object. To train this model, we propose a benchmark driver gaze dataset, UD-FSG (Urban Driving-Face Scene Gaze), comprising synchronized driver-face and traffic-scene images, bounding boxes for scene objects, and gaze labels expressed as 2D gaze coordinates and corresponding gaze objects. We introduce a hybrid loss function comprising classification, consistency, confusion-aware attention, and hard-negative margin losses to improve the robustness of the proposed TransGaze-Object model during training. 
\par  The proposed TransGaze-Object model achieves an overall gaze-object prediction accuracy of 60\%. In comparison, associating the predicted gaze point with object bounding boxes using a state-of-the-art PoG-based approach achieves an overall accuracy of 51\%. Thus, the proposed model provides an approximately 9\% improvement in gaze-object prediction accuracy, corresponding to a 17.5\% relative improvement over SGAP-Gaze, highlighting the benefit of incorporating object-level geometric and spatial information during training to establish a more reliable association between driver gaze and scene objects. The error analysis shows that TransGaze-Object reduces confusion between traffic objects and the background, with an error rate of 11.68\%, compared with 23.21\% for the PoG-based gaze-object association using post-processing. However, distinguishing between traffic objects and the background remains challenging. Further error analysis shows that gaze prediction on smaller objects is challenging. By formulating driver gaze estimation as an end-to-end gaze object prediction problem, this work aims to stimulate further research into object-level representations of driver visual attention.
\par In future work, the proposed framework can be further improved in terms of prediction accuracy, robustness, and generalization across diverse driving conditions. The framework can be extended to support different camera configurations and dashboard-mounted camera placements by explicitly incorporating the corresponding camera intrinsic and extrinsic parameters. Furthermore, the current framework can be expanded beyond the forward scene to include side windows and a wider portion of the vehicle interior, enabling driver gaze estimation over a broader field of view. Future studies can also investigate temporal information from consecutive frames to model the dynamic nature of driver gaze and improve the stability of gaze-object predictions under challenging real-world driving conditions.

\section*{Declaration of Generative AI and AI-assisted technologies in the writing process}
The authors declare that ChatGPT was used to assist with grammar checks and language corrections in this manuscript. All content generated by the tool was carefully reviewed, revised, and approved by the authors, who take full responsibility for the content of the publication.

\section*{CRediT authorship contribution statement}
% \textbf{Pavan Kumar Sharma:} Conceptualization, Formal analysis, Investigation, Methodology, Writing-Original Draft.
% \textbf{Pranamesh Chakraborty:} Conceptualization, Funding acquisition, Investigation, Methodology, Supervision, Writing-Review \& Editing.
\textbf{Pavan Kumar Sharma:} Conceptualization, Data curation, Formal analysis, Investigation, Methodology,  Software, Validation, Writing-original draft.
\textbf{Ayush Pande:} Conceptualization, Investigation, Writing-Review \& Editing.
\textbf{Pranamesh Chakraborty:} Conceptualization, Investigation, Methodology,  Resources, Supervision, Validation, Writing-Review \& Editing.

\section*{Declaration of competing interest}
The authors declare that they have no known competing financial interests or personal relationships that could have appeared to influence the work reported in this paper.

\section*{Data availability}
The dataset supporting the findings of this study is  available at the following link: 
\url{https://github.com/pavans20/Urban-Driving-Face-Scene-Gaze-Dataset.git}.

%\section*{Generative AI and Figures, images and other artwork}

% \section*{Acknowledgement}
% Our research results are based upon work supported by the Initiation Grant scheme of Indian Institute of Technology Kanpur (IITK/CE/2019378). Any opinions, findings, and conclusions or recommendations expressed in this material are those of the author(s) and do not necessarily reflect the views of the IITK.

\bibliography{bibliography}
\end{document}